\documentclass{cambrian}
\usepackage[utf8]{inputenc} 
\usepackage[T1]{fontenc}    
\usepackage{hyperref}       
\usepackage{url}            
\usepackage{booktabs}       
\usepackage{amsfonts}       
\usepackage{nicefrac}       
\usepackage{microtype}      

\usepackage{mwe}
\usepackage{graphicx}
\usepackage{amssymb}
\usepackage{tabularray}
\usepackage{adjustbox}
\usepackage{colortbl}
\usepackage{array}
\usepackage{multirow}
\usepackage{makecell}
\usepackage{bbm}
\usepackage{collcell,xfp}
\usepackage{pgf}
\usepackage{tikz}
\usepackage[most]{tcolorbox}
\usepackage{csquotes}
\usepackage[noorphans,vskip=1em,leftmargin=1em]{quoting}
\usepackage{pifont} 
\usepackage{enumitem} 
\usepackage{tcolorbox} 
\usepackage{forest}
\usepackage{wrapfig}

\usepackage{caption}
\usepackage{subcaption}
\usepackage{longtable}
\usepackage{algpseudocode}
\usepackage{amsmath}
\usepackage[ruled,vlined,linesnumbered]{algorithm2e}
\usepackage[capitalize]{cleveref}
\crefname{algocf}{Algorithm}{Algorithms}
\Crefname{algocf}{Algorithm}{Algorithms}
\usepackage[percent]{overpic}
\usepackage{xspace}
\usepackage[normalem]{ulem}  
\usepackage{tocloft}  
\usepackage{natbib}  
\usepackage{mathtools}
\usepackage{listings}

\usepackage{enumitem} 
\usepackage{multirow}
\usepackage{amssymb}
\usepackage{wasysym}
\usepackage{bbm}
\usepackage{soul}
\usepackage{fontawesome}

\newcommand{\huggingface}{\raisebox{-1.5pt}{\includegraphics[height=1.05em]{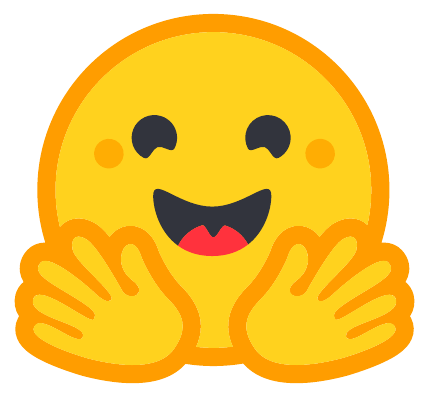}}\xspace}
\newcommand{\github}{\raisebox{-1.5pt}{\includegraphics[height=1.05em]{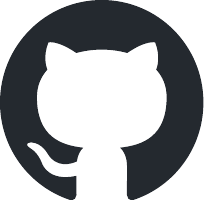}}\xspace}
\newcommand{\worldwideweb}{\raisebox{-1.5pt}{\includegraphics[height=1.05em]{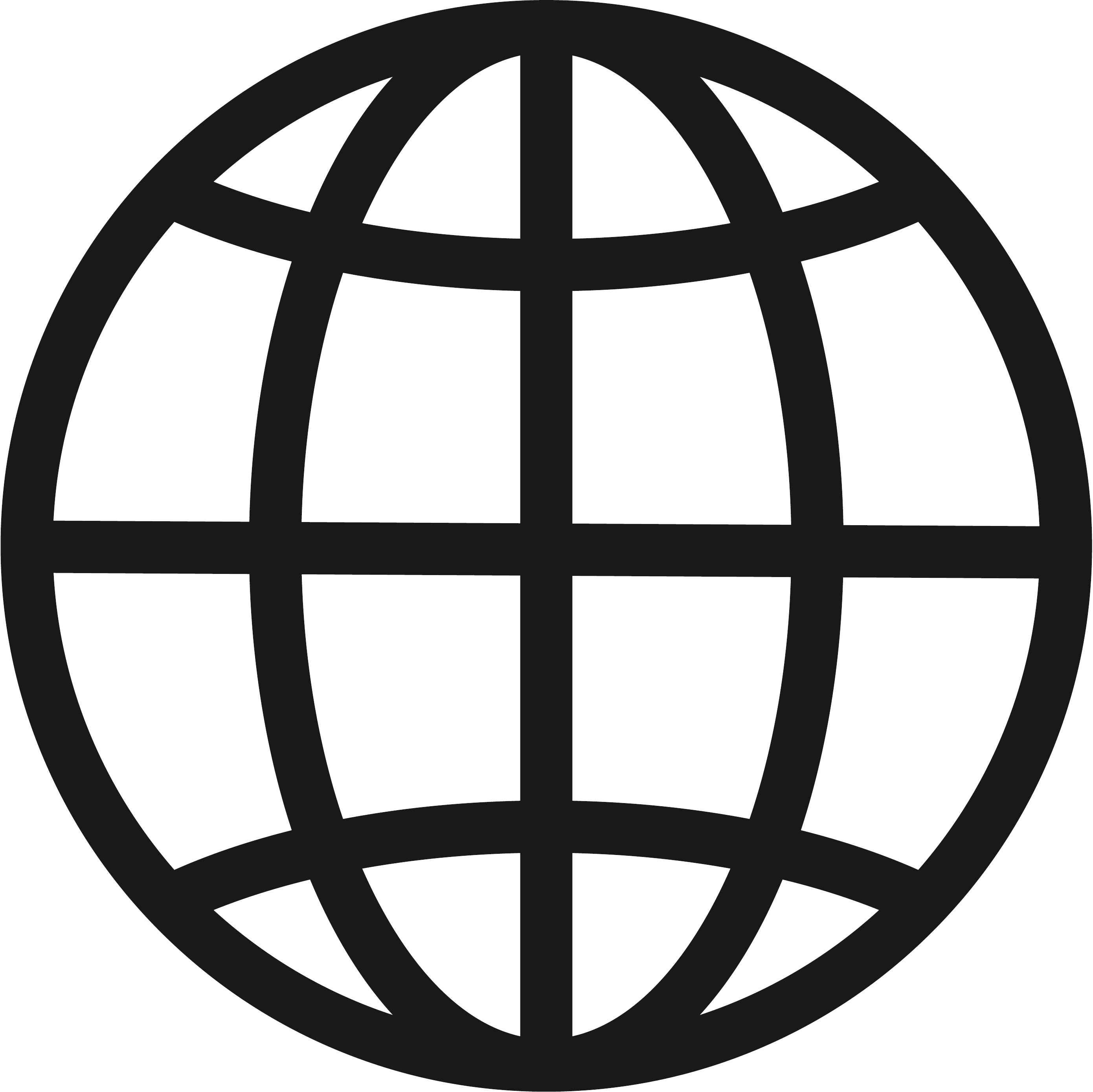}}\xspace}

\definecolor{scholarblue}{rgb}{0.21,0.49,0.74}
\definecolor{bluelink}{RGB}{0,113,188}
\definecolor{greenlink}{RGB}{0,188,113}
\hypersetup{
    colorlinks=true,%
    citecolor=scholarblue,%
    filecolor=red,%
    linkcolor=red!93!black,%
    urlcolor=bluelink
}

\usepackage{titlesec}
\titlespacing*{\paragraph}
    {0pt}     
    {0.25em}  
    {1em}  

\newcommand{\finding}[2]{
    \begin{tcolorbox}[
        colback=white!90!gray,     
        colframe=teal!60!black,     
        arc=5pt,                    
        boxsep=5pt,                 
        left=10pt,                  
        right=10pt,                 
        top=2pt,                    
        bottom=2pt,                 
        boxrule=0.8pt,              
        drop shadow=gray!50!white,  
        enhanced jigsaw             
    ]
    #2
    \end{tcolorbox}
}

\definecolor{navyblue}{HTML}{0071BC}
\iftrue   
\newcommand{\displaytodo}[1]{#1}
\else
\newcommand{\displaytodo}[1]{}
\fi

\def\eg{\emph{e.g.}} 
\def\ie{\emph{i.e.}}

\definecolor{diffgreen}{RGB}{213, 245, 213}
\definecolor{diffred}{RGB}{255, 215, 215}

\definecolor{diffgreen}{RGB}{213, 245, 213}
\definecolor{diffred}{RGB}{255, 215, 215}

\definecolor{blindcolor}{HTML}{AB2AC6}    
\definecolor{chancecolor}{HTML}{F59E0B}   
\definecolor{singlecolor}{HTML}{06B6D4}   
\definecolor{multiplecolor}{HTML}{2563EB} 
\definecolor{captioncolor}{HTML}{22C55E}  

 \newcommand{\culine}[2]{%
    \def\temp@uline{\bgroup\markoverwith
        {\textcolor{#1}{\rule[-0.5ex]{2pt}{1pt}}}\ULon}%
    \temp@uline{#2}%
}
 \newcommand{\cthickuline}[3][0.8pt]{%
    \def\temp@uline{\bgroup\markoverwith
        {\textcolor{#2}{\rule[-0.5ex]{2pt}{#1}}}\ULon}%
    \temp@uline{#3}%
}

\title{\center{Benchmarking Visual State Tracking \\ in Multimodal Video Understanding}}

\renewcommand\Affilfont{\normalfont\fontsize{11}{15}\selectfont\centering}
\author{
    Sihyun~Yu$^{1,2\dagger *}$\quad
    Nanye~Ma$^{1\dagger *}$\quad
    Pinzhi~Huang$^{1\dagger *}$\quad
    Hyunseok~Lee$^{2*}$\quad
    Shusheng~Yang$^{1}$\\
    June~Suk~Choi$^{2}$\quad 
    Ellis~Brown$^{1}$\quad
    Oscar~Michel$^{1}$\quad 
    Boyang~Zheng$^{1}$\quad 
    Jinwoo~Shin$^{2}$\quad 
    Saining~Xie$^{1}$\\
    $^{1}$New York University\quad
    $^{2}$KAIST
}

\begingroup

\footnotetext{$^\dagger$ Project lead.}
\footnotetext{$^*$ Equal technical contribution.}
\endgroup

\tcbuselibrary{skins, breakable} 
\definecolor{cornellred}{rgb}{0.7, 0.11, 0.11}
\definecolor{cadmiumgreen}{rgb}{0.0, 0.42, 0.24}
\definecolor{aliceblue}{rgb}{0.91, 0.94, 0.97}
\definecolor{darkblue}{rgb}{0.83, 0.89, 0.97}
\definecolor{Blue9}{rgb}{0.098,0.3,0.9}
\definecolor{mutedviolet}{HTML}{EEDAFF} 
\definecolor{mutedpink}{HTML}{FFD9D9} 
\definecolor{mutedblue}{HTML}{CCE5FF} 
\definecolor{mutedgreen}{HTML}{CEF8C2} 
\definecolor{forestgreen}{rgb}{0.13, 0.55, 0.13}
\definecolor{hookergreen}{rgb}{0.0, 0.44, 0.0}
\definecolor{cozygreen}{RGB}{170, 210, 160} 
\definecolor{vviolet}{HTML}{676aa4} 
\definecolor{vpink}{HTML}{bb7699} 
\definecolor{vblue}{HTML}{618eb4} 
\definecolor{vgreen}{HTML}{367660} 
\definecolor{vgreen}{HTML}{367660} 
\definecolor{bananamania}{rgb}{0.98, 0.91, 0.71}

\definecolor{navyblue}{HTML}{0071BC}
\definecolor{hotpink}{HTML}{FF0080}

\definecolor{oai-white}{HTML}{FFFFFF}
\definecolor{oai-black}{HTML}{000000}
\definecolor{oai-red}{HTML}{FF4500}
\definecolor{oai-green}{HTML}{51DA4C}
\definecolor{oai-blue}{HTML}{0000FF}
\definecolor{oai-yellow}{HTML}{FFF639}
\definecolor{oai-magenta}{HTML}{FF45FF}
\definecolor{oai-cyan}{HTML}{00FFFF}
\definecolor{oai-orange}{HTML}{FE7600}
\definecolor{oai-violet}{HTML}{8A2BE2}
\definecolor{oai-brown}{HTML}{A0522D}
\definecolor{oai-green-050}{HTML}{F4FFF4}
\definecolor{oai-green-100}{HTML}{E9FFE8}
\definecolor{oai-green-200}{HTML}{D9FFD8}
\definecolor{oai-green-300}{HTML}{C9FFC7}
\definecolor{oai-green-400}{HTML}{A6FFA3}
\definecolor{oai-green-500}{HTML}{7CF178}
\definecolor{oai-green-600}{HTML}{51DA4C}
\definecolor{oai-green-700}{HTML}{3FA93B}
\definecolor{oai-green-800}{HTML}{2D712A}
\definecolor{oai-green-900}{HTML}{193718}
\definecolor{oai-gray-000}{HTML}{FFFFFF}
\definecolor{oai-gray-100}{HTML}{FAFAFA}
\definecolor{oai-gray-200}{HTML}{F5F5F5}
\definecolor{oai-gray-300}{HTML}{E5E5E5}
\definecolor{oai-gray-400}{HTML}{FFB7A4}
\definecolor{oai-gray-500}{HTML}{CDCDCD}
\definecolor{oai-gray-600}{HTML}{A8A8A8}
\definecolor{oai-gray-700}{HTML}{747474}
\definecolor{oai-gray-800}{HTML}{393939}
\definecolor{oai-gray-900}{HTML}{000000}

\definecolor{tablesectionbg}{HTML}{FAF4F8} 
\definecolor{stateelementtable}{HTML}{E7CCEF}   
\definecolor{statestructuretable}{HTML}{F3C6DA} 
\definecolor{softteal}{HTML}{74C9BD}
\definecolor{softpeach}{HTML}{E99A7A}

\newcommand{\textft}[1]{#1}
\newcommand{\customul}[2][black]{\setulcolor{#1}\setul{0.25ex}{1.5pt}\ul{#2}\setulcolor{black}}

\definecolor{statefaint}{HTML}{F0F0F0}
\newcommand{\statedimfaint}[1]{{\sethlcolor{statefaint}\hl{\textbf{#1}}}}

\definecolor{elem}{HTML}{9E35BE}
\definecolor{elembg}{HTML}{9E35BE}
\newcommand{\elemdim}[1]{{\sethlcolor{elem!25}\hl{\textbf{#1}}}}
\newcommand{\elemcode}[1]{{\textft{\customul[elem]{#1}}}}             

\definecolor{struct}{HTML}{D9578F}
\definecolor{structbg}{HTML}{F3C6DA}
\newcommand{\structdim}[1]{{\sethlcolor{structbg!90}\hl{\textbf{#1}}}}
\newcommand{\structcode}[1]{{\textft{\customul[struct]{#1}}}}

\definecolor{perception}{HTML}{5EC26A}
\definecolor{perceptionbg}{HTML}{5EC26A}
\newcommand{\percpdim}[1]{{\sethlcolor{perceptionbg!35}\hl{\textbf{#1}}}}
\newcommand{\percpcode}[1]{{\textft{\customul[perception]{#1}}}}

\newcommand{\tabledarkhl}[1]{{\sethlcolor{oai-gray-500}\hl{#1}}}
\newcommand{\tablelighthl}[1]{{\sethlcolor{oai-gray-300}\hl{#1}}}

\hypersetup{
  linkcolor = cornellred,
  citecolor  = cadmiumgreen,
  colorlinks = true,
  urlcolor = Blue9
}

\newcommand{\task}{visual state tracking\xspace}
\newcommand{\sname}{\textsc{VSTAT}\xspace}

\newcommand*{\ShowNotes}{} 

\ifdefined\ShowNotes
  \newcommand{\colornote}[3]{{\color{#1}\bf{#2: #3}\normalfont}}
\else
  \newcommand{\colornote}[3]{}
\fi

\def\eg{\emph{e.g.}} 
\def\ie{\emph{i.e.}}

\newcommand{\pmark}{\textcolor{black!45}{\LEFTcircle}}
\newcommand{\totalvid}{834\xspace}%
\newcommand{\totalq}{1,500\xspace}%

\newcommand{\GOOD}{{\color{ggreen}\ding{51}}}
\newcommand{\BAD}{{\color{gred}\ding{55}}}

\definecolor{ggreen}{rgb}{0.53, 0.69, 0.43}
\definecolor{gred}{HTML}{F5433D}

\def\rvq{{\mathbf{q}}}

\def\rvv{{\mathbf{v}}}

\def\rvy{{\mathbf{y}}}

\begin{abstract}
Understanding a video requires more than recognizing isolated moments, as humans continuously track entities, states, and events over time. This capacity for \emph{visual state tracking} is fundamental to video understanding, yet remains underexplored in current evaluations of Multimodal Large Language Models (MLLMs). We introduce \textbf{V}isual \textbf{STA}te \textbf{T}racking benchmark (VSTAT), a video-based benchmark designed to diagnose visual state tracking in MLLMs. VSTAT consists of 834 clips drawn from both synthetic and real-world videos, paired with 1,500 questions that cannot be answered from any single frame or short segment, requiring continuous perception and integration of events across the entire video stream. Despite their strong performance on existing video benchmarks, we find that state-of-the-art MLLMs perform far below humans and only modestly above answer-prior baselines. To analyze this gap, we compare MLLMs' thinking traces with the underlying video stream to understand \emph{why} and \emph{when} MLLMs fail on VSTAT. We find that MLLMs reason and track correctly in text, but fail at visually perceiving the events they need to track.
Finally, our preliminary evaluation suggests that recent agentic approaches, including MLLM-based video agents and coding agents, do not readily resolve these failures, still falling short on VSTAT.    
\end{abstract}

\begin{document}

\maketitle


\begin{center}
    \renewcommand{\arraystretch}{1.5}
    \begin{tabular}{rll}
        \worldwideweb{} & \textbf{Website} & \url{https://vision-x-nyu.github.io/vstat-site/}\\
        \huggingface{} & \textbf{Benchmark} & \url{https://huggingface.co/collections/nyu-visionx/vstat} \\
        \github{} & \textbf{Evaluation code} & \url{https://github.com/vision-x-nyu/vstat}\\
    \end{tabular}
\end{center}


\newpage
{
    \hypersetup{linkcolor=black}
    \tableofcontents
}
\newpage

\section{Introduction}
\label{sec:intro}
Videos are not just a discrete sequence of RGB pixels; they are records of continuous dynamics and processes in the visual world \citep{yang2026cambrians}. When we watch a video, we do not simply perceive each frame independently, but also understand and analyze the underlying dynamics by keeping track of essential information. For instance, when watching a basketball game, we naturally keep track of the score and who attempted each shot by making sense of complex visual procedures. This capacity for \emph{\task} is fundamental to how humans learn from and reason about visual demonstrations.

Recent Multimodal Large Language Models (MLLMs) have progressed remarkably in video understanding, demonstrating strong capabilities in semantic understanding and action recognition \citep{Bai2025Qwen3VLTR, yang2026cambrians, Wang2025InternVL35AO, google2026gemini31pro}. However, it remains unclear whether current MLLMs can understand continuous dynamics and track evolving states throughout a procedure presented in the video \citep{Wang2024ContinuousPM}, which is essential for real-world applications such as robotics \citep{black2410pi0}. This gap stems from the fact that existing video understanding benchmarks are mostly not explicitly designed for evaluating this capability. In many cases, the answer can be inferred by relying on a small subset of keyframes, salient moments, or visible end states, without continuously tracking how the underlying state evolves over time. As a result, strong performance on these benchmarks does not necessarily indicate an ability to track necessary information in the video. While a few recent works have attempted to address this gap \citep{Liu2026CanVM, Wang2024ContinuousPM}, their evaluations remain limited to a single synthetic task (\eg, shell game) and do not cover diverse, real-world scenarios.

\begin{figure*}[ht!]
\centering\small
\vspace{-0.3in}
\includegraphics[width=.93\textwidth]{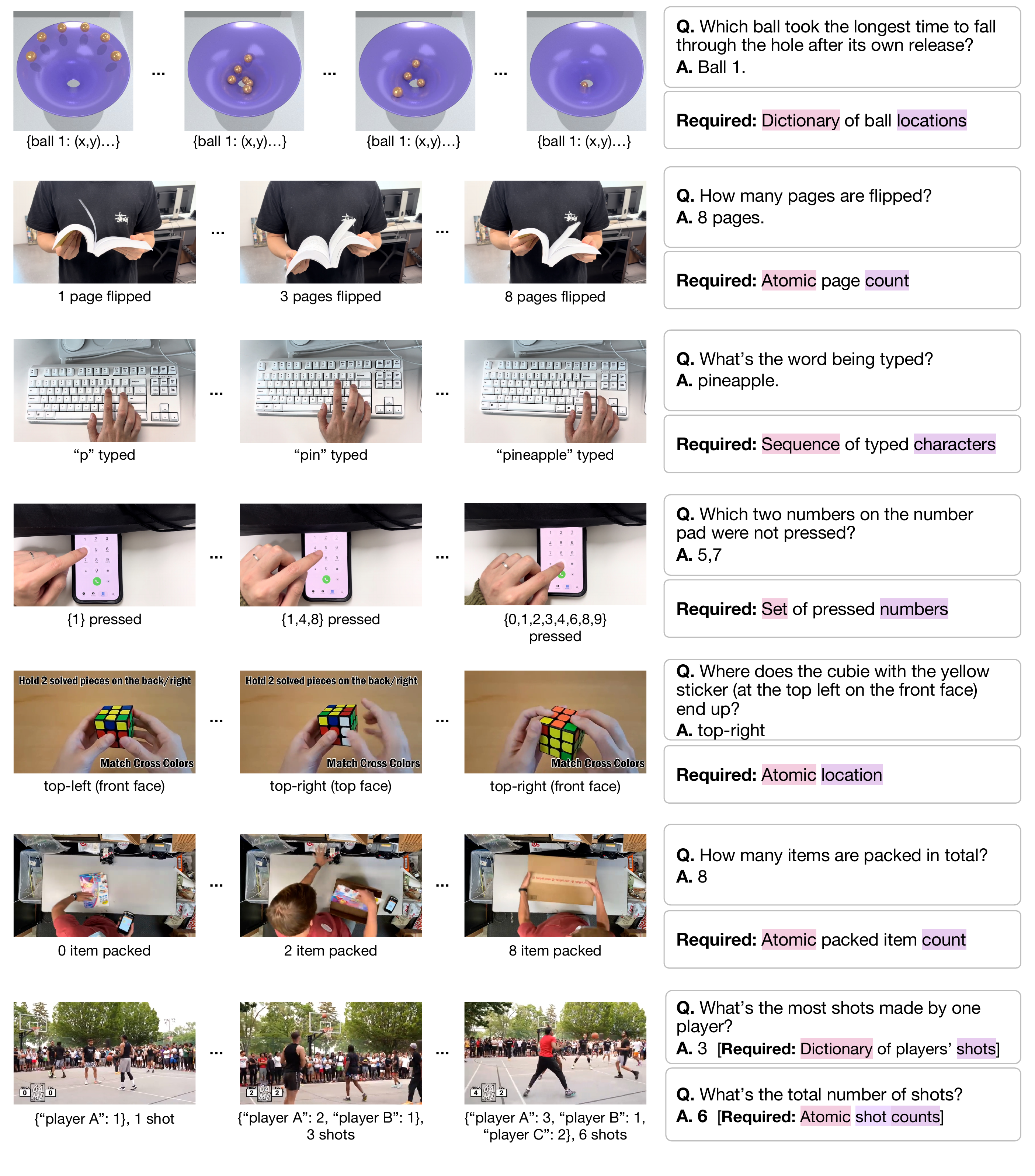}
\setlength{\abovecaptionskip}{-9pt}
\caption{
\textbf{Task examples in \sname.} 
All questions require \task to answer. For illustration, we simplified the questions and subsampled video frames. Each example requires tracking different states, which are combinations of \structdim{structure} and \elemdim{element type}.
}
\label{fig:gallery}
\vspace{-0.1in}
\end{figure*}
We introduce \textbf{V}isual \textbf{STA}te \textbf{T}racking benchmark (\sname), a video-based benchmark designed to diagnose \task in MLLMs.\footnote{We note that our formulation of \task extends beyond object tracking in pixel space: our benchmark also covers tracking of underlying latent state representations in the video stream.} \sname adopts a standard question-answering format, where the model receives a video stream and a question as input and outputs an answer. \sname consists of \totalvid~video clips paired with \totalq~questions drawn from synthetic, self-recorded, and real-world videos in the wild that contain procedural processes. Each task is constructed so that the answer cannot be read off any single keyframe or a few salient moments: critical events may be hidden, visually similar to each other, or distributed across multiple entities and moments. Thus, models must continuously perceive and integrate events throughout the entire video stream. The tasks in \sname vary in the complexity of state to be tracked, and exhibit various perceptual challenges in extracting state from video; for instance, in the Rubik's cube task, the model must track a specific cubie even when it is occluded in some frames~(see Figure~\ref{fig:gallery} for more examples). Surprisingly, while these questions can be easily answered by humans, state-of-the-art MLLMs perform far below humans and only modestly above answer-prior baselines.

To understand this gap, we investigate behaviors of MLLMs through several analyses. Firstly, we study \emph{why} MLLMs struggle to solve \sname by conducting controlled experiments on synthetic tasks generated in Blender environments. We first test whether frame subsampling, which MLLMs usually apply to videos and may cause them to miss brief events, is the bottleneck. To do so, we compare performance on the original videos against temporally stretched versions, where each event spans more frames to ensure that subsampling does not introduce ambiguity. However, we observe only marginal improvement, suggesting that this is not the case. 

Then, to further investigate whether the failures of MLLMs stem from insufficient reasoning or limited perception capabilities, we use several simple tasks in \sname whose underlying events can be manually transcribed. We compare model performance under two conditions: the original video input and a text transcription that explicitly describes each frame and event (see Figure~\ref{subfig:transcription}). While MLLMs struggle in the video condition, they solve the same tasks almost perfectly when given text transcriptions. This contrast suggests that the fundamental bottleneck lies in visual perception of the video stream, rather than in flawed reasoning. We stress, however, that such transcription is infeasible for most tasks in \sname due to their more complex settings and the substantially larger amount of information required to answer the questions (see Figure~\ref{fig:gallery}). Ultimately, tackling the state-tracking challenge requires much stronger perceptual capabilities from MLLMs.

\begin{table}[t]
\vspace{-0.25in}
\centering\small
\caption{
\textbf{Comparison with existing video understanding benchmarks.} 
We compare \sname against existing benchmarks in terms of their coverage of state-tracking tasks. \pmark~denotes that the benchmark contains some instances satisfying the categories but only as a small fraction. CP-Bench and VET-Bench focus on 1--2 simple synthetic tasks (counting identical cubes and shell game, respectively). Dataset sources: Scripted (C), Real (R) and Synthetic (Y). 
}

\begin{tabular}{lcccccc}
\toprule
Benchmark & Source & \#Clips & \#QAs & State Tracking & Real-world & Diverse \\
\midrule
VideoMME-v2 \citep{Fu2026VideoMMEv2TT} & C,R & 800 & 3{,}200 & \pmark & \GOOD & \GOOD \\
VideoReasonBench \citep{liu2026videoreasonbench} & C,Y & 240 & 1{,}440 & \GOOD & \pmark & \BAD \\
CP-Bench \citep{Wang2024ContinuousPM} & Y & 101 & 101 & \GOOD & \BAD & \BAD \\
VET-Bench \citep{Liu2026CanVM} & Y & 100 & 100 & \GOOD & \BAD & \BAD \\
\midrule
\textbf{\sname (ours)} & C,R,Y & \totalvid & \totalq & \textbf{\GOOD} & \textbf{\GOOD} & \textbf{\GOOD} \\
\bottomrule
\end{tabular}

\label{tab:comparison}
\end{table}

Secondly, we diagnose \emph{when} MLLMs fail during \task by analyzing mismatches between their textual thinking traces and the input video stream in failure cases. From their traces, we identify three major failure modes: event recognition, entity association (\ie, linking the same entity across frames), and state update (\ie, updating the tracked state after each perceived event). For instance, when watching a shell game video, the model may misidentify which cups were swapped, lose track of the cup hiding a target item, or fail to update its location even after correctly identifying it. Even with recent agentic frameworks, including MLLM-based video agents \citep{wang2025active} and state-of-the-art coding agents \citep{anthropic2026opus47, singh2025openai}, these failures cannot be readily mitigated and the performance gap remains substantial.

\noindent
We highlight the main contributions of this paper below:
\vspace{-0.1in}
\begin{itemize}[leftmargin=*,itemsep=0mm]
\item We introduce \sname, a video-based benchmark for evaluating the \emph{\task} capability of MLLMs, covering both synthetic and real-world videos paired with questions.
\item We show that state-of-the-art MLLMs perform far below human performance and only modestly above answer-prior baselines on \sname.
\item Through controlled experiments and analyses, we find that perceiving task-relevant events from the continuous {visual} stream is a major bottleneck.
\item  We demonstrate that recent agentic frameworks, including video agent methods and coding agents, do not improve the performance on our benchmark.
\end{itemize}

\vspace{-0.02in}
\section{\sname: Visual State Tracking Benchmark}
\vspace{-0.02in}
Our benchmark, \sname, is designed to evaluate the \emph{\task} capability of MLLMs throughout a continuous video stream. \sname follows the standard format of video benchmarks for MLLMs: given a video stream $\rvv$ and a query $\rvq$, the model $f$ must predict the answer $\rvy$. Unlike prior video MLLM benchmarks, we construct \sname such that the answer cannot be inferred from a single keyframe or a small subset of frames. 
Instead, every task in \sname requires the model to process the entire video stream, track and update the information needed to derive the answer. 
One of the most popular examples is the shell game \citep{Liu2026CanVM}, which tests a player's observational skills by having them follow a hidden object as three cups are shuffled; the player must maintain the target cup's location throughout the video.

\sname comprises a diverse set of tasks requiring \task capabilities, drawn from both synthetic and real-world videos. Concretely, \sname consists of \totalvid video clips paired with \totalq questions, derived from simulated videos rendered with Blender and real-world videos collected from YouTube and our own recordings. \sname covers diverse tasks with varied tracking targets, such as counting packed items, recognizing typed words, or attributing shots to players. This enables extensive evaluation and analysis of the \task capability of models across diverse video streams that contain continuous procedural processes. We provide illustrative examples in Figure~\ref{fig:gallery}, along with dataset statistics and a comparison with existing video benchmarks in Table~\ref{tab:comparison}.

\subsection{Data curation}
\label{subsec:curation}
In the rest of this section, we refer to the task examples illustrated in Figure~\ref{fig:gallery}.

\vspace{0.02in}\noindent
\textbf{Video curation.}
We curate our videos from both simulated environments and the real world. For simulated videos, we design 9 environments using the 3D software Blender and synthesize 450 video clips in total. For real-world videos, we collect 304 video clips from YouTube and record 80 additional videos ourselves in scripted settings. As a result, \sname contains \totalvid~video clips in total. Across both sources, we focus on videos that contain diverse procedural processes such as solving puzzles, athletic plays, cooking, and order packing. 
In addition, each clip contains factors that make \percpdim{perception} difficult; for example, the basketball clip involves continuous \percpcode{camera movement}, while the order-packing clip exhibits frequent \percpcode{occlusion} between items. We provide details of video categories, preprocessing strategies, and additional example visualizations in Appendix~\ref{appen:data}.

\vspace{0.02in}\noindent
\textbf{Question-answer generation.}
From each video clip, we design questions that require \task over the video stream to predict an answer. We follow two design principles. First, all questions are designed to avoid visual shortcuts: the answer cannot be inferred from a few keyframes or the visible end state, forcing models to track state throughout the video. For instance, the book example's ``how many pages are flipped?'' cannot be answered without tracking the entire video. Second, we use diverse query types that require tracking of various \elemdim{element types} and their \structdim{structures}. Some queries require tracking \elemcode{locations} (\eg, ``\emph{where} does the cube with the yellow sticker end up?''), while others require tracking total \elemcode{counts} (\eg, ``what is the \emph{total number} of shots?'') or \elemcode{attributes} such as characters (\eg, ``what is the \emph{word} being typed?''). Each query also demands a different state structure: it can be \structcode{atomic} when tracking a single position or counter, or more complex structures such as \structcode{sequences} or \structcode{sets} when the query asks about detailed history of the video stream.

\finding{}{\faBookmark~~\sname questions cannot be solved by visual shortcuts such as sparse keyframes or the end state, requiring models to maintain and update state across the entire video.}

We design multiple questions for each video, with each question requiring the model to track different types of information. As illustrated by the basketball example, both the amount of information needed to answer a question and the associated difficulty can vary substantially depending on the query. Consequently, our benchmark contains \totalq~questions in total. We believe this “one video, multiple questions” format enables a comprehensive analysis of different aspects of models’ visual state tracking capabilities.
Our questions come in two formats: numerical questions (NQs), whose answers are single numbers, and multiple-choice questions (MCQs), which are used for all other question types and include carefully designed distractors. All videos, questions, answers, and category labels are annotated and reviewed through a human-in-the-loop verification protocol; see Appendix~\ref{appen:data} for details.

\subsection{Taxonomy}
As explained in Section~\ref{subsec:curation}, our benchmark involves two crucial complementary axes: \emph{perceptual complexity}, which captures factors in the video stream that make visual perception difficult, and \emph{state complexity}, which captures the amount and type of minimum information that must be extracted from the video to answer the question. In what follows, we explain in detail the categories we define along each axis to classify each instance in our benchmark.

\vspace{0.02in}\noindent
\statedimfaint{State complexity.}
As shown in the examples of Figure~\ref{fig:gallery}, each instance requires a different state complexity, which we decompose into two orthogonal dimensions: \elemdim{element type} and \structdim{structure}. We consider three categories for \elemdim{element type}: \elemcode{count} (book example), \elemcode{location} (cube example), and \elemcode{attribute} (keyboard example). For \structdim{structure}, we consider four categories: \structcode{atomic} (book example), \structcode{sequence} (keyboard example), \structcode{set} (numberpad example), and \structcode{dictionary} (basketball example).

\vspace{0.02in}\noindent
\percpdim{Perceptual complexity.}
From the video and question-answer pairs we collected, we consider the following six categories related to factors that make video perception difficult: \percpcode{occlusion} (\eg, the cube is hidden in some frames), \percpcode{camera motion} (\eg, the camera moves and the scene changes in the basketball clip), \percpcode{homogeneity} (\eg, multiple cubes share similar appearances), \percpcode{symbolic decoding} (\eg, typing events must be transcribed into characters), \percpcode{multi-entity attribution} (multiple players move simultaneously in the basketball clip), and \percpcode{event ambiguity} (\eg, page flips can occur in either direction in the book example). Note that these categories reflect the major axes we observed across diverse procedural tasks, and may be extended to capture further variations. 

With this taxonomy, we label each video-question pair along all three axes and use these labels to ensure a balanced data distribution for the benchmark that is not skewed toward any particular aspect;  in Appendix~\ref{appen:data}, we provide the detailed statistics breakdown of our benchmark, including the aforementioned axes, duration of each video clip, and keywords in the questions. For future research, we also open-source these labels along with the questions and video clips in our benchmark.

\begin{table*}[t!]
    \centering
    \caption{\textbf{Evaluation on \sname.} Scores report the reparsed MRA-with-MCQ metric. \tabledarkhl{Dark gray} indicates the best result among all models and \tablelighthl{light gray} indicates the best result among open-sourced models. Ranks are computed separately within proprietary API models (1--4) and within open-sourced models (1--20, pooling Thinking and Instruct); baselines are not ranked.}
\begin{minipage}[c]{0.59\textwidth}
\vspace{0pt}
    \footnotesize                          
    \setlength\tabcolsep{3pt}%
    \renewcommand{\arraystretch}{1.15}
\resizebox{\linewidth}{!}{
\begin{tabular}{r|cc|ccc|cccc}
 & & &
 \rotatebox{55}{Count} &
 \rotatebox{55}{Location} &
 \rotatebox{55}{Attribute} &
 \rotatebox{55}{Atomic} &
 \rotatebox{55}{Sequence} &
 \rotatebox{55}{Set} &
 \rotatebox{55}{Dict} \\
Methods & Rank & Avg. & \multicolumn{3}{c|}{\cellcolor{stateelementtable}State Element} & \multicolumn{4}{c}{\cellcolor{statestructuretable}State Structure} \\
\hline
\rowcolor{tablesectionbg}
\multicolumn{1}{l|}{\textcolor{black}{\textit{Baselines}}} & & & & & & & & & \\
Chance Level (Random)    & - & 26.1 & 25.0 & 26.7 & 25.0 & 28.2 & 25.0 & 25.0 & 25.0 \\
Chance Level (Frequency) & - & 37.8 & 41.3 & 33.5 & 35.1 & 39.2 & 26.6 & 41.4 & 39.9 \\
Human Performance             & - & 90.5 & 92.8 & 89.9 & 86.4 & 93.7 & 77.5 & 90.0 & 92.4 \\
\hline
\rowcolor{tablesectionbg}
\multicolumn{1}{l|}{\textcolor{black}{\textit{Proprietary Models (API)}}} & & & & & & & & & \\
Gemini-3.1 Pro (low) \citep{google2026gemini31pro}   & \cellcolor{oai-green-200}{1} & \cellcolor{oai-gray-500}{44.4} & \cellcolor{oai-gray-500}{42.6} & 38.5 & \cellcolor{oai-gray-500}{54.1} & 39.5 & 60.8 & \cellcolor{oai-gray-500}{51.9} & 38.7 \\
Gemini-3.1 Pro (high) \citep{google2026gemini31pro}   & \cellcolor{oai-green-100}{2} & 43.9 & 42.1 & 41.6 & 49.9 & \cellcolor{oai-gray-500}{40.1} & 56.8 & 50.0 & \cellcolor{oai-gray-500}{39.3} \\
Gemini-3.0 Flash (low) \citep{google2025gemini3flash} & 3 & 39.8 & 33.4 & 40.3 & 52.2 & 32.5 & 61.6 & 48.2 & 35.2 \\
Gemini-3.0 Flash (high) \citep{google2025gemini3flash} & 4 & 38.8 & 33.2 & 36.6 & 52.5 & 31.4 & \cellcolor{oai-gray-500}{62.4} & 48.4 & 32.4 \\
\hline
\rowcolor{tablesectionbg}
\multicolumn{1}{l|}{\textcolor{black}{\textit{Open-sourced Models Thinking}}} & & & & & & & & & \\
MiMo-VL-7B \citep{Yue2025MiMoVLTR}                     & 11 & 31.2 & 28.3 & 32.6 & 35.7 & 26.9 & \cellcolor{oai-gray-300}{40.0} & 33.8 & \cellcolor{oai-gray-300}{32.8} \\
InternVL3.5-8B-Thinking \citep{Wang2025InternVL35AO}    & 13 & 30.2 & 24.5 & 32.6 & 39.5 & 26.0 & 35.5 & 41.4 & 27.6 \\
GLM-4.1V-9B-Thinking \citep{zhipu2025glm41v}            & 14 & 30.2 & 24.8 & 33.9 & 37.2 & 26.9 & 33.2 & 40.8 & 27.3 \\
Qwen3VL-8B-Thinking \citep{Bai2025Qwen3VLTR}           & 18 & 28.2 & 25.9 & 32.2 & 28.6 & 26.8 & 32.7 & 28.5 & 28.0 \\
Qwen3VL-4B-Thinking \citep{Bai2025Qwen3VLTR}           & 19 & 26.0 & 21.0 & 31.8 & 30.2 & 25.5 & 30.8 & 29.5 & 21.4 \\
\hline
\rowcolor{tablesectionbg}
\multicolumn{1}{l|}{\textcolor{black}{\textit{Open-sourced Models Instruct}}} & & & & & & & & & \\
LLaVA-OV-2-8B \citep{llavaonevision2}         & \cellcolor{oai-green-200}{1} & \cellcolor{oai-gray-300}{35.1} & 28.3 & \cellcolor{oai-gray-500}{43.0} & 40.5 & 33.5 & 38.7 & \cellcolor{oai-gray-300}{46.9} & 27.3 \\
LLaVA-OV-2-8B (codec) \citep{llavaonevision2}          & \cellcolor{oai-green-100}{2} & 35.0 & 28.6 & 42.0 & \cellcolor{oai-gray-300}{40.6} & 33.9 & 37.0 & 46.3 & 27.6 \\
Molmo2-4B \citep{molmo2}                               & 3 & 34.4 & 31.6 & 39.7 & 34.5 & \cellcolor{oai-gray-300}{37.1} & 33.6 & 36.7 & 27.1 \\
Cambrian-S-7B \citep{yang2026cambrians}                & 4 & 34.2 & \cellcolor{oai-gray-300}{33.2} & 33.6 & 36.9 & 34.0 & 30.6 & 40.2 & 32.5 \\
Molmo2-8B \citep{molmo2}                               & 5 & 34.0 & 30.9 & 37.0 & 37.0 & 34.7 & 36.3 & 39.1 & 27.0 \\
Qwen3VL-8B \citep{Bai2025Qwen3VLTR}                    & 6 & 33.2 & 30.9 & 37.0 & 33.9 & 32.4 & 33.3 & 37.9 & 31.5 \\
InternVL3.5-2B \citep{Wang2025InternVL35AO}            & 7 & 31.8 & 29.6 & 33.9 & 34.1 & 31.7 & 29.9 & 36.3 & 29.9 \\
Cambrian-S-3B \citep{yang2026cambrians}                & 8 & 31.8 & 29.7 & 32.7 & 35.0 & 32.7 & 31.9 & 35.1 & 27.2 \\
VITA-1.5-7B \citep{fu2025vita15}                       & 9 & 31.5 & 25.5 & 36.3 & 38.6 & 29.4 & 33.0 & 43.1 & 26.3 \\
Qwen3VL-4B \citep{Bai2025Qwen3VLTR}                    & 10 & 31.3 & 27.0 & 33.3 & 37.9 & 30.4 & 32.8 & 39.8 & 25.8 \\
InternVL3.5-8B \citep{Wang2025InternVL35AO}            & 12 & 30.6 & 25.1 & 33.2 & 39.2 & 26.9 & 33.8 & 41.8 & 28.3 \\
Qwen3VL-2B \citep{Bai2025Qwen3VLTR}                    & 15 & 29.4 & 29.4 & 28.2 & 30.5 & 32.5 & 24.9 & 32.1 & 23.5 \\
Cambrian-S-1.5B \citep{yang2026cambrians}              & 16 & 29.3 & 26.0 & 34.1 & 31.0 & 28.0 & 31.0 & 31.8 & 29.3 \\
LLaVA-OV-7B \citep{li2024llava}                        & 17 & 28.6 & 20.1 & 34.8 & 39.4 & 24.5 & 30.0 & 43.8 & 25.0 \\
LLaVA-OV-0.5B \citep{li2024llava}                      & 20 & 21.3 & 14.6 & 33.9 & 21.7 & 19.7 & 25.8 & 22.2 & 20.9 \\
\end{tabular}
 }
\end{minipage}\hfill
\begin{minipage}[c]{0.39\textwidth}
\vspace{0pt}
\centering
\includegraphics[width=\linewidth,trim=8pt 7pt 8pt 7pt,clip]{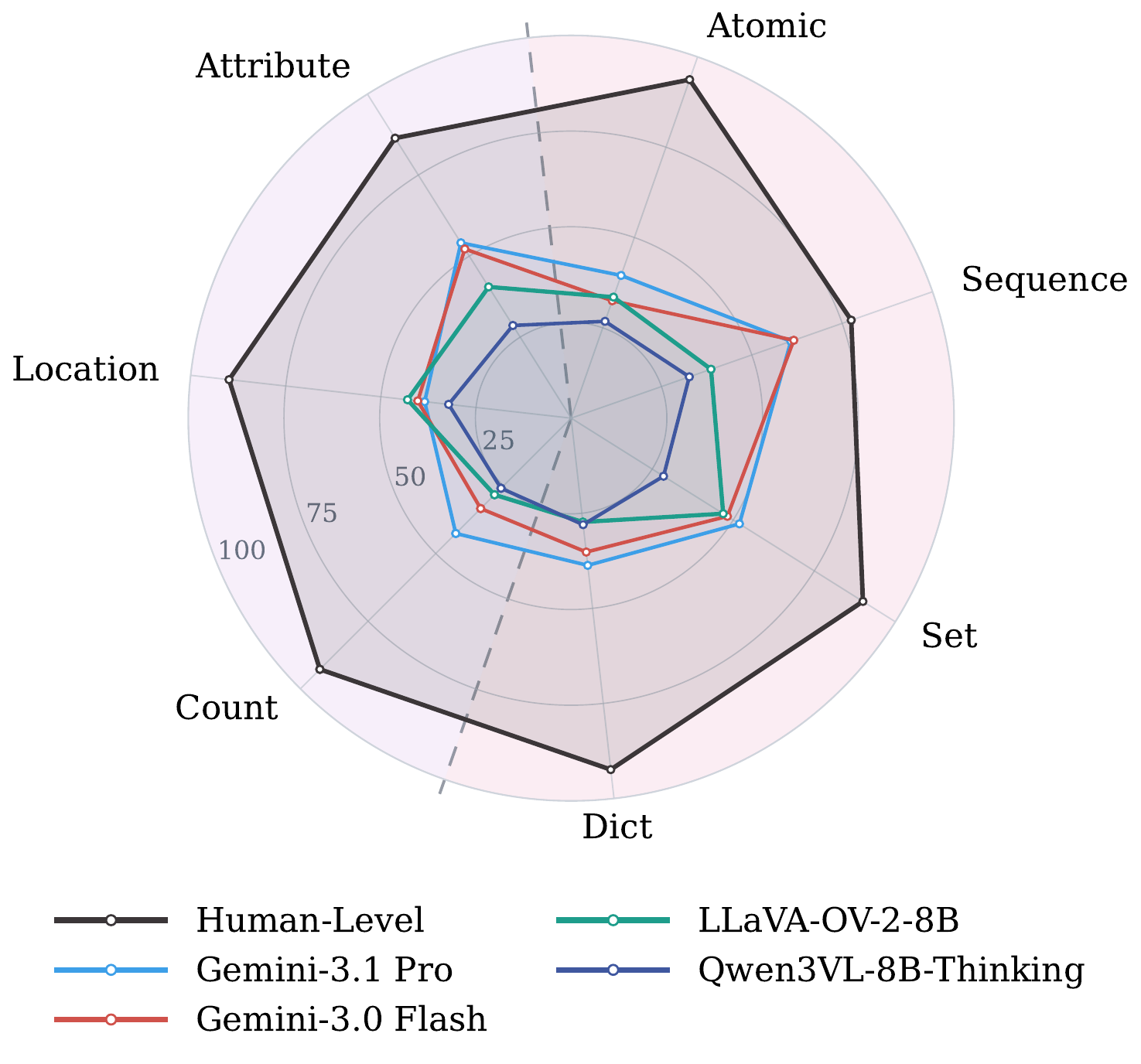}
\end{minipage}
\label{tab:main}
\vspace{0.03in}
\begin{minipage}{0.98\textwidth}
\vspace{0.1in}
\small\emph{Note.} 
\textit{Each question is labeled by state element (Count, Location, Attribute) and state structure (Atomic, Sequence, Set, Dict). Avg. is computed over all questions, not as the mean of bucket scores. }
\end{minipage}
\end{table*}

\section{Evaluation on \sname}
Using \sname, we evaluate the performance of recent MLLMs and their shortcomings, including (a) proprietary models (API; Gemini-3.1 Pro \citep{google2026gemini31pro} and Gemini 3.0 Flash \citep{google2025gemini3flash}) with different thinking levels, and open-sourced models (Qwen3VL \citep{Bai2025Qwen3VLTR}, Cambrian-S \citep{yang2026cambrians}, MiMo-VL \citep{Yue2025MiMoVLTR}, InternVL3.5 \citep{Wang2025InternVL35AO}, LLaVA-OV \citep{li2024llava}, LLaVA-OV-2 \citep{llavaonevision2}, and Molmo2 \citep{molmo2}), by varying their model sizes and thinking mode, if applicable. We also consider several agentic frameworks, both specialized for video understanding (AVP \citep{wang2025active}) and coding (Claude Code \citep{anthropic2026opus47} and Codex \citep{singh2025openai}) for our studies. 

\noindent
In particular, we investigate the following questions:
\vspace{-0.1in}
\begin{itemize}[leftmargin=*]
\item How well do state-of-the-art MLLMs perform on \sname overall? (Table~\ref{tab:main}, Table~\ref{tab:thinking_performance_scores})
\item Why do MLLMs fail to solve tasks in \sname? (Table~\ref{tab:blender_stretch}, Figure~\ref{fig:pilot})
\item When do MLLMs fail to solve tasks in \sname? (Figure~\ref{fig:failure})
\item Can recent agentic frameworks solve tasks in \sname? (Table~\ref{tab:agentic})
\end{itemize}

\subsection{Setup}
\label{subsec:setup}

\textbf{Metrics and evaluation protocol.}
Our evaluation pipeline builds on LMMs-Eval~\citep{zhang2024lmmseval} and follows the standard evaluation protocol of MLLMs on video benchmarks.
Following VSI-Bench \citep{yang2024think}, we report the average of accuracy on MCQs and mean relative accuracy (MRA) on NQs. For open-sourced models, we sweep the maximum frame budget over $\{16, 32, 64, 128\}$ uniformly sampled frames and report the best score for each model; the selected budgets are 32 frames for Qwen3VL-8B \citep{Bai2025Qwen3VLTR} and Cambrian-S-7B \citep{yang2026cambrians}, 64 frames for Qwen3VL-4B, Qwen3VL-2B, and LLaVA-OV-2-8B \citep{llavaonevision2}, 128 frames for Molmo2-8B \citep{molmo2}, and 16 frames for all other models. We additionally report LLaVA-OV-2-8B with its codec video backend, which packs codec-sampled frames into canvases (32 canvases from up to 256 sampled frames) instead of uniform frame sampling. 
For proprietary models (Gemini~
\citep{google2026gemini31pro}), we set the resolution parameters as MEDIUM for evaluation, as we observe no significant performance difference across resolution parameters, and set $\mathtt{max}\_\mathtt{tokens}$=65536 during evaluation for sufficient reasoning budget.

\vspace{0.02in}
\noindent
\textbf{Chance level baselines.}
Following VSI-Bench, we provide two baselines: Chance Level (Random) is the random selection accuracy for MCQ tasks (and is inapplicable for NQ tasks). Chance Level (Frequency) represents the highest performance MLLMs would achieve by always selecting the most frequent answer for each task. This identifies performance gains that may result from inherently long-tailed answers or imbalanced multiple-choice distributions. We also report human performance as a sanity check, measured by authors who were not involved in constructing the corresponding questions, which shows the difficulty of \sname for humans. See Appendix~\ref{appen:eval} for more details.

\subsection{Main results}
In Table~\ref{tab:main}, we report the evaluation results across three state \elemdim{elements} (\elemcode{count}, \elemcode{location}, and \elemcode{attribute}) and four state \structdim{structures} (\structcode{atomic}, \structcode{sequence}, \structcode{set}, \structcode{dictionary}), along with the overall average accuracy and rank. As shown in the table, only Gemini-3.1 Pro and Gemini-3.0 Flash are modestly above the Chance-Level (Frequency) answer-prior baseline, while other models perform even worse. In contrast, humans solve our benchmark with an average accuracy of 90.5\%, far exceeding the chance-level baseline and existing MLLMs. This reveals that a large gap still exists between the \task capabilities of humans and MLLMs. One exception is tasks that require tracking of \structcode{sequence} states, which show opposite trends between humans and MLLMs: for humans this is the most challenging category compared with other state structures, but for MLLMs it is the best-performing category, showing the smallest gap. We also observe that all open-sourced models perform worse than Chance-Level (Frequency) across all numbers of frames fed into the models, and usually show only marginal improvement with increased model size, with Molmo2 and InternVL3.5 demonstrating slight degradation.

Notably, although LLaVA-OV-2~\citep{llavaonevision2} and Molmo2~\citep{molmo2} are specifically trained with motion-grounded codec streams and pixel-space object tracking data, respectively, they do not demonstrate substantial improvements over other open-source MLLMs, despite being the two best-performing open-source models. This further suggests that \sname evaluates a more complex form of state tracking that goes beyond pixel-level tracking or motion-grounding objectives, requiring models to track the underlying latent state representations evolving throughout the video stream.

\begin{wraptable}{r}{0.54\linewidth}
\vspace{-0.95em}
\centering
\footnotesize
\setlength{\tabcolsep}{4.5pt}
\renewcommand{\arraystretch}{1.18}
\begin{tabular}{@{}l l c c@{}}
\toprule
\textbf{Model} & \textbf{Thinking} & \textbf{Performance} & \textbf{$\Delta$} \\
\midrule
Gemini-3.1-Pro 
& low $\to$ high
& 44.4 $\to$ 43.9
& \textcolor{red}{$-1.1\%$} \\
Gemini-3.0-Flash
& low $\to$ high
& 39.8 $\to$ 38.8
& \textcolor{red}{$-2.5\%$} \\
Qwen3VL-8B
& w/o $\to$ w/
& 33.2 $\to$ 28.2
& \textcolor{red}{$-15.1\%$} \\
InternVL3.5-8B
& w/o $\to$ w/
& 30.6 $\to$ 30.2
& \textcolor{red}{$-1.3\%$} \\
\bottomrule
\end{tabular}
\vspace{-0.4em}
\caption{\textbf{Thinking does not reliably improve performance.} $\Delta$ reports the relative performance change.}
\label{tab:thinking_performance_scores}
\vspace{-0.9em}
\end{wraptable}

We also observe that enabling thinking mode or increasing thinking levels hurts performance, as shown in Table~\ref{tab:thinking_performance_scores}. Gemini-3.1-Pro is only mildly affected by higher thinking levels, with performance changing from $44.4$ to $43.9$, while Gemini-3.0-Flash drops from $39.8$ to $38.8$. Among open-source models, Qwen3VL-8B exhibits a substantial decline from $33.2$ to $28.2$, whereas InternVL3.5-8B shows only slight degradation, moving from $30.6$ to $30.2$. Notably, this observation aligns with the findings of~\cite{Fu2026VideoMMEv2TT}. After inspecting examples in Appendix~\ref{appen:think_failures}, we find that, for tasks with higher perceptual complexity, a larger thinking budget can increase the likelihood of hallucination for these models.

\finding{}{\faBookmark~~\sname is generally solvable by humans, but existing MLLMs struggle to solve it.}



\begin{figure*}[t!]
\captionsetup[subfigure]{justification=centering}
\centering\small
\begin{subfigure}[t]{.55\textwidth}
\centering
\includegraphics[height=4.1cm]{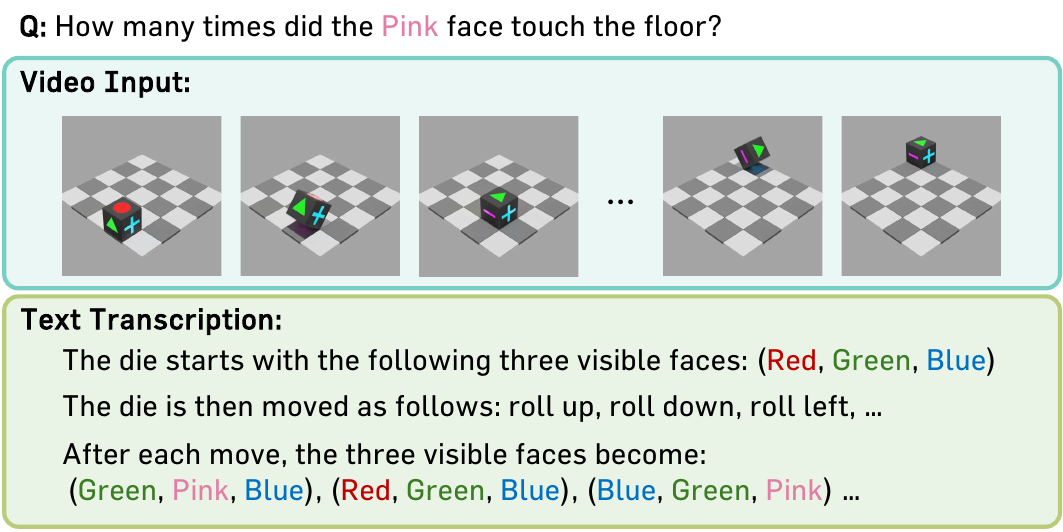}
\caption{Example task and its text transcription. } 
\label{subfig:transcription}
\end{subfigure}
\begin{subfigure}[t]{.43\textwidth}
\centering
\includegraphics[height=4.1cm]{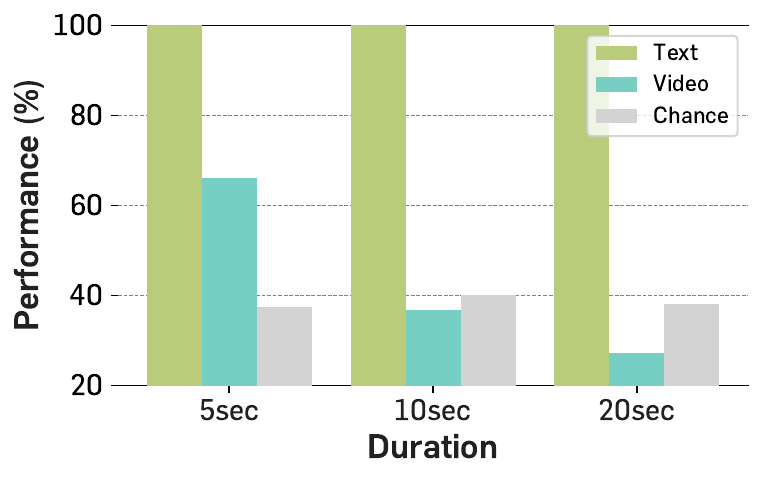}
\caption{Performance across video durations.} 
\label{subfig:tv_gap}
\end{subfigure}
\caption{\textbf{Analyzing bottlenecks of MLLMs in \task.} 
(a) An example Blender task (rolling die) with its video frames and text transcription. 
(b) Performance across video durations on the selected task subset. Recent MLLMs, such as Gemini-3.1 Pro \citep{google2026gemini31pro}, solve the task perfectly with text conditions, but their video performance drops to near chance and degrades further as videos grow longer.}
\label{fig:pilot}
\end{figure*}

\subsection{\emph{Why} do current MLLMs fail to solve \sname?}
To analyze why this large gap occurs, we conduct additional experiments from two different perspectives. First, we examine whether this performance gap stems from event ambiguity caused by the information loss from frame subsampling when feeding the video to the model. Second, we investigate whether the gap arises from the model's visual perception or its reasoning capability. To this end, we take several simple tasks from the Blender environment, which allow us to control the number of events and the video duration, making them suitable for controlled analysis.

\begin{wraptable}{r}{0.35\linewidth}
    \vspace{-0.1cm}
    \centering\small
        \begin{tabular}{ll}
            \toprule
                Data & Avg. \\
                \midrule
                Chance level (Freq.) & 39.2 \\
                \cmidrule(lr){1-2}
                5sec & 51.4 \\
                5sec + stretch & 53.6 \\
                \bottomrule
        \end{tabular}
    \vspace{-0.05cm}
        \caption{\textbf{Impact of video stretching}, evaluated on Gemini-3.1 Pro.}
        \label{tab:blender_stretch}
    \vspace{-0.3cm}
\end{wraptable} 

\vspace{0.02in}\noindent
\textbf{This gap does not mainly stem from event ambiguity.}
To rule out potential event ambiguity caused by the model's relatively low video sampling rate, we first compare the performance of Gemini-3.1 Pro on the original 5-second Blender videos with that on their temporally stretched versions, where each original frame is duplicated five times. This ensures that every event in the video is fully visible to Gemini even under the 1 FPS sampling rate.\footnote{Every event in the Blender videos lasts at least 0.2s; after temporal stretching, each event spans at least one second, making it fully visible to the model.} However, as shown in Table~\ref{tab:blender_stretch}, performance only marginally improves, suggesting that event ambiguity from frame subsampling is not the primary cause of the gap, which instead reflects fundamental limitations in the model's visual perception capability.

\vspace{0.02in}\noindent
\textbf{This gap stems from visual perception.}
We conduct an additional experiment to disentangle whether MLLM failures on \task tasks stem from visual perception or from reasoning. Specifically, we compare model performance under two conditions: the original video, and a text-only counterpart in which the video is replaced by a textual transcription of the visible states and events. If the gap between the two conditions is large, it suggests that visual perception, rather than reasoning, could be the primary bottleneck. Specifically, we consider three simple Blender tasks whose visible observations and events can be easily transcribed into text. For example, we consider the rolling die task, which requires counting how often a specific face lands on the bottom. Here, the transcription describes the three visible faces and the rolling direction at each step (see Figure~\ref{subfig:transcription}).

As shown in Figure~\ref{subfig:tv_gap}, state-of-the-art MLLMs solve these tasks \emph{perfectly} when given textual transcriptions, yet their performance on video inputs drops to random-guess level once the video exceeds 10 seconds. Crucially, even on 5-second videos, where context length is negligible, performance already falls considerably short of the perfect text-only accuracy. While longer videos further degrade performance, the gap is already substantial at 5 seconds, suggesting that visual perception is the primary bottleneck, with errors compounding over longer videos. We observe the same pattern on other simple Blender-synthesized tasks (see Appendix~\ref{appen:transcription}).

\finding{}{\faBookmark~~Visual perception appears to be a main bottleneck for models on \sname.}
Here, we emphasize that text transcription is not intended as a solution, but rather as a diagnostic tool for probing the fundamental bottleneck behind MLLMs’ failures on \sname.
On these Blender tasks, the perceptual gap is already so severe that we had to provide the text transcriptions by hand: even state-of-the-art MLLMs fail to reliably transcribe these simple synthetic videos into text.
For real-world videos containing more complex dynamics and richer visual details (see Figure~\ref{fig:gallery} and additional examples in Appendix~\ref{appen:subsec:examples}), text transcription would be even more challenging; in many cases, the resulting descriptions can exceed the length of the videos themselves, making this approach infeasible.

\begin{figure}[t!]
    \centering
    \includegraphics[width=\textwidth]{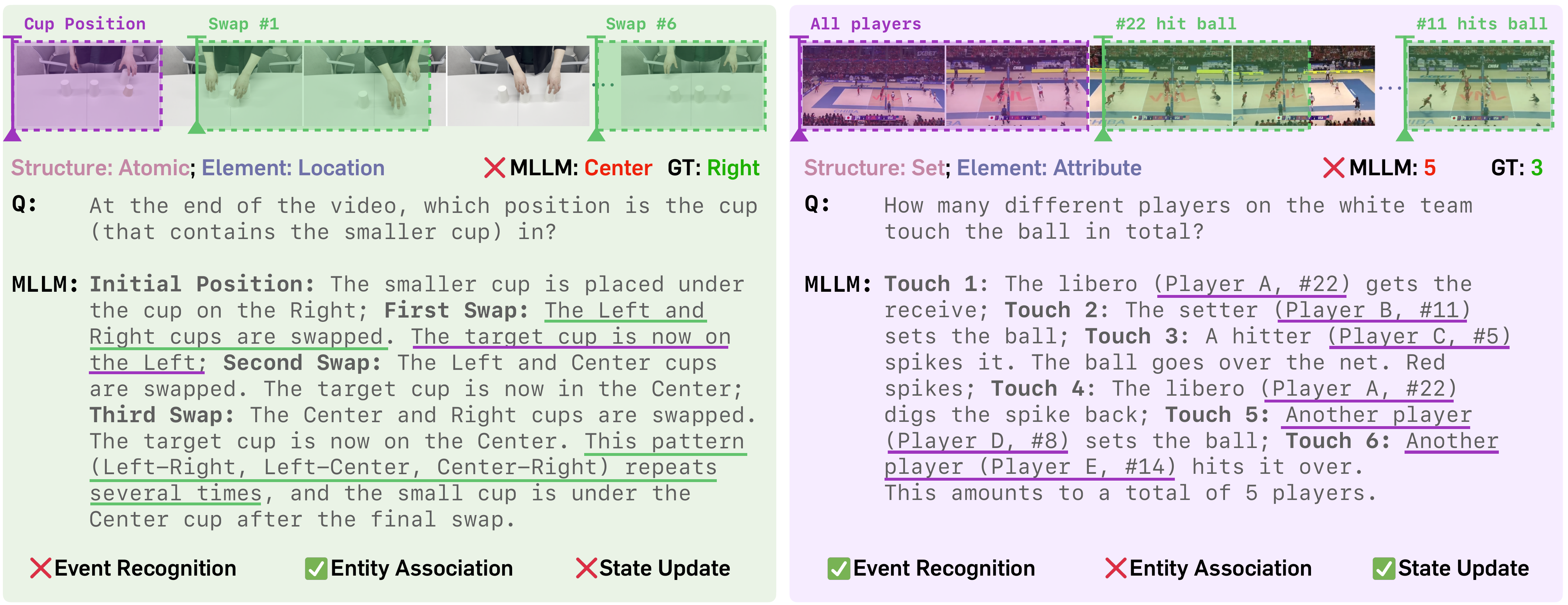}
    \caption{\textbf{Failures in event identification.} We highlight phrases and frames related to state extraction in \elemcode{purple} and failures in visual perception in \percpcode{green}. For better illustration, we subsampled video frames related to the failures and simplified the thinking traces.}    
    \label{fig:failure}
\end{figure}

\subsection{\emph{When} do current MLLMs fail to solve \sname?}
Next, we examine when MLLMs fail by analyzing the thinking traces of Gemini-3.1 Pro, the best-performing model in our benchmark. Comparing each video with its trace, we identify three recurring failure modes, summarized in Figure~\ref{fig:failure}. We also conduct a quantitative error analysis, as shown in Figure~\ref{fig:error_analysis}. More examples and details are in Appendix~\ref{appen:failure}.

\noindent
\textbf{Event recognition.} Even for relatively straightforward events in the video, the model can fail to correctly recognize the event and extract the corresponding state information. In the left example of Figure~\ref{fig:failure}, the person swaps the center and right cups, but the model identifies the event as “the Left and Right cups are swapped,” leading to an incorrect cup location and ultimately an incorrect final answer. In more challenging cases, we observe that the model may even hallucinate the entire event trace without correctly identifying any of the actual events (see Appendix~\ref{appen:failure}).

\begin{wrapfigure}{r}{0.5\textwidth}
\vspace{-0.25in}
    \includegraphics[width=\linewidth]{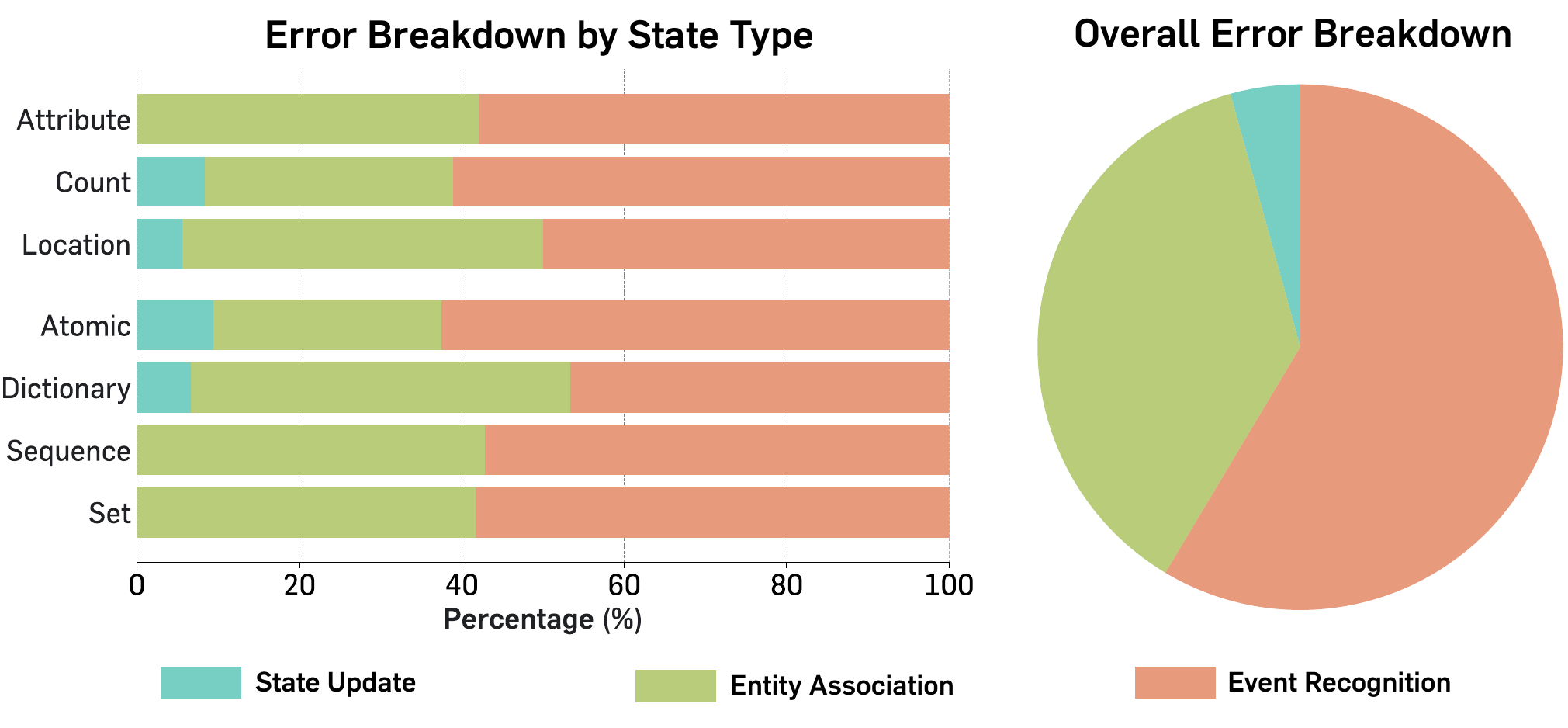}
    \setlength{\abovecaptionskip}{-4pt}
    \caption{\textbf{Human-conducted error analysis.} More than $50\%$ are from event recognition.}
    \label{fig:error_analysis}
    \vspace{-0.2in}
\end{wrapfigure}

\noindent
\textbf{Entity association.} Besides misidentifying events in the video, the model often fails when the state requires consistent association with a specific entity among visually similar objects. In the volleyball example, all players wear identical uniforms, so distinguishing them requires motion-based tracking. While the model correctly identifies each ball-touch event, it assigns a new random jersey number each time the ball is touched, even when the same player handles the ball repeatedly and their actual number becomes visible later.

\noindent
\textbf{State update.} Lastly, we observe an interesting failure pattern in which the model correctly recognizes the events and/or maintains the correct entity associations throughout the video, but fails to use this information to update the state needed for question answering. For example, in the left example of Figure~\ref{fig:failure}, during the third swap, the model correctly identifies that ``the Center and Right cups are swapped'' and that the target cup was previously in the Center. However, it incorrectly updates the target cup’s location as remaining in the Center, whereas it should move to the Right. We observe that this occurs more often when the model needs to track continuous trajectories, as it tends to over-simplify the observations and loses significant information from the video stream.

\noindent
\textbf{Quantitative analysis.}
As summarized in Figure~\ref{fig:error_analysis}, most failures are due to errors in either event recognition or entity association. In particular, more than 50\% of failures stem from event recognition, suggesting that the dominant bottleneck of current MLLMs may lie in low-level perception, rather than visual reasoning. We also observe that state update errors highly correlate with the model’s textual reasoning capability. Therefore, this error type remains relatively limited, likely due to the advanced reasoning abilities of current state-of-the-art MLLMs.

\subsection{Can agentic frameworks improve performance on \sname?}

Finally, we conduct a preliminary case study examining whether recent agentic frameworks built on (M)LLMs can achieve better performance on \sname. We consider the following three agentic frameworks: AVP \citep{wang2025active}, a recent video agent; Codex with GPT-5 \citep{singh2025openai} and Claude Code with Opus 4.7 \citep{anthropic2026opus47}, two state-of-the-art coding agents. For Codex and Claude Code, we provide the video file directory and the corresponding question, and ask the model to write visual reasoning code to solve the question.\footnote{Here, we find that the coding agents, Codex and Claude Code, are prone to contamination, so we carefully rule out potential answer leakage; see Appendix~\ref{appen:agentic} for details and examples.} We report results on a small subset of \sname in Table~\ref{tab:agentic} by selecting one clip per video category.

\begin{wraptable}{r}{0.36\linewidth}
    \vspace{-0.1in}
    \resizebox{\linewidth}{!}{
    \centering\small
        \begin{tabular}{lc}
            \toprule
                Method & Avg. \\
                \midrule
                Chance level (Freq., $n$=39 subset) & 50.8 \\
                \cmidrule(lr){1-2}
                Gemini-3.1 Pro \citep{google2026gemini31pro} & 52.6 \\
                Gemini-3.1 Pro \citep{google2026gemini31pro} $+$ AVP \citep{wang2025active} & 43.6 \\
                Claude Code (Opus 4.7, max) \citep{anthropic2026opus47}   &37.6 \\
                Codex (GPT-5, xhigh) \citep{singh2025openai}   &53.4 \\
                \bottomrule
        \end{tabular}
        }
        \caption{\textbf{Agentic method results.}}
        \label{tab:agentic}
    \vspace{-0.4cm}
\end{wraptable} 
As shown in the table, agentic methods are not able to solve \sname; rather, they show near chance-level accuracy despite their strong performance on text-based tasks, further indicating that the primary bottleneck of solving VSTAT lies in the visual perception capabilities of current models. We also observe that coding agents typically spend considerable time and tokens to answer a question. Solving a single question takes approximately 30 minutes on average, largely because they produce inconsistent intermediate results in their thinking traces, which confuses the model itself, even resulting in a wrong answer. For video agentic frameworks, we observe that they tend to show the opposite mode: they commit too early to their initial observation, sampling the video at a fixed low frame-rate (typically 1 FPS) and synthesizing an answer from a single round of evidence collection without verification. We include evaluation details and examples in Appendix~\ref{appen:agentic}.

\vspace{-0.03in}
\section{Related Work}
\vspace{-0.03in}

\noindent
\textbf{Video Multimodal Large Language Models (MLLMs).}
Recent progress in multimodal understanding largely stems from MLLMs \citep{tong2024cambrian, yang2026cambrians, bai2023qwenvl, wang2024qwen2vl, Bai2025Qwen3VLTR}, which incorporate powerful foundational visual encoders \citep{tschannen2025siglip, radford2021learning, Simeoni2025DINOv3} into the strong linguistic understanding capabilities of LLMs \citep{brown2020language, bai2023qwen, touvron2023llama2}. This success in the image domain has naturally led to the exploration 
of video-based MLLMs \cite{li2024llama, li2024llava, zhang2024video, song2024moviechat, bai2025qwen2, zhu2025internvl3, yang2026cambrians, comanici2025gemini, google2026gemini31pro, Bai2025Qwen3VLTR}, which is essential for real-world applications that require multimodal intelligence, such as robotics \citep{black2410pi0, bjorck2025gr00t} and web agents \citep{gou2025navigating}. Our benchmark evaluates and diagnoses the \task capabilities of MLLMs, which are essential for many applications such as long-horizon robotic manipulation.

\noindent
\textbf{Evaluation of video MLLMs.}
To effectively measure progress, pinpoint current limitations, and guide future research, a series of benchmarks have been proposed to evaluate video MLLMs from different perspectives, including general video understanding~\citep{fu2025video, li2024mvbench, Fu2026VideoMMEv2TT}, event recognition~\citep{xiao2021next, caba2015activitynet}, knowledge reasoning~\citep{hu2025video, zhao2025mmvu}, and temporal grounding and reasoning~\citep{gao2017tall, cai2024temporalbench, shangguan2024tomato}.
More recent efforts impose stricter requirements, challenging models to comprehend hours- or even day-long videos~\citep{song2024moviechat, mangalam2023egoschema, chandrasegaran2024hourvideo, wu2024longvideobench, wang2024lvbench, zhou2024mlvu, wang2023lifelongmemory} and to reason about the spatial information underlying video frames~\citep{yang2024think, yang2026cambrians}.
Despite the breadth of these efforts, little to none attention has been paid to \textit{visual state tracking}---the ability to continuously monitor visual states and events as they evolve over time. This capability is effortless for humans yet indispensable for real-world applications such as robotic manipulation, assistive agents, and surveillance, and it remains conspicuously absent from existing evaluation suites. This paper aims to fill this gap by proposing a benchmark to evaluate and diagnose \task capability of MLLMs. 

\noindent
\textbf{Comparison with concurrent works.}
\sname is related to several concurrent benchmarks, including VET-bench 
\citep{Liu2026CanVM}, VideoReasonBench~\citep{liu2026videoreasonbench} and Video-MME-v2 \citep{Fu2026VideoMMEv2TT}, but differs substantially in scope and design. VET-bench shares our motivation of evaluating state tracking in MLLMs, but is limited to two shell-game-like tasks in a simulated environment with only 100 video clips---an order of magnitude smaller than \sname. Video-MME-v2 is a comprehensive video understanding benchmark that includes a few categories relevant to state tracking (\eg, repetitive action counting and entity persistence tracking).
In contrast, \sname systematically covers tracking of underlying latent state representations in the video stream. Finally, videos in VideoReasonBench are either synthetic or recorded under scripted setups, and many videos explicitly visualize the events (\eg, swaps shown as arrows), introducing visual shortcuts. In contrast, \sname contains real-world videos with no explicit visual cues for the underlying events.

\noindent
\textbf{Video world models.}
Our benchmark shares some conceptual similarity with video world models \citep{gao2026dreamdojo, lecun2022path, ha2018recurrent, hafner2023mastering, guo2025mineworld, sun2025worldplay, kanervisto2025world, hong2025relic, team2026advancing, kong20253d, alonso2024diffusion, genie3}, which aim to predict future states from previous states, actions, and observations. The main difference is that these methods typically assume actions are given explicitly and define the state representation as an approximation of the entire visual world, usually represented as latent video representations \citep{lecun2022path} or the entire sequence of video frames, including predicted ones \citep{gao2026dreamdojo}. In contrast, our setting assumes actions are implicitly given through events, and the state is defined relative to the query, capturing only the partial information from the video necessary to answer it. We hope this connection also facilitates better evaluation of world modeling.

\section{Conclusion}
We present \sname, a video-based benchmark for diagnosing the \task capability of MLLMs. Our evaluation reveals a substantial gap between humans and current MLLMs, which only modestly exceed answer-prior baselines. Through controlled analyses, we further identify visual perception, rather than textual tracking, as the primary bottleneck, and diagnose recurring failure modes. Finally, we show that existing agentic frameworks, including video agents and coding agents, do not trivially resolve these failures. We hope \sname serves as a useful diagnostic tool for the community to understand and improve the visual perception of MLLMs on continuous, real-world video streams. We discuss limitations and future directions in Appendix~\ref{appen:limitation}.

\section*{Acknowledgments}
We thank Taeyoung Kim, Anjali Gupta, and Ying Wang for proofreading, and thank Daohan Lu for helping with our human evaluation. S.X. acknowledges support from the MSIT IITP grant (RS-2024-00457882) and NSF Award IIS-2443404.

\addcontentsline{toc}{section}{References}
\bibliography{main}
\bibliographystyle{plain}


\clearpage
\appendix
\clearpage

\section{Benchmark Breakdown}
\label{appen:data}
\subsection{Detailed Information}

\textbf{Formal definition}
In Table~\ref{tab:state_types} and \ref{tab:challenges}, we provide a more formal definition of each category in our taxonomy, which is used for our labeling process (detailed in Appendix~\ref{appen:subsec:curation}).

\begin{table}[ht!]
\caption{\textbf{Taxonomy of state structures} across diverse queries and tasks (Table \ref{tab:simulated_videos} and~\ref{tab:real_videos}).}
\centering \small
\begin{tabular}{l p{5.5cm} p{7cm}}
\toprule
\textbf{Type} & \textbf{Definition} & \textbf{Examples} \\
\midrule
\multicolumn{3}{l}{\emph{Element type}} \\
\addlinespace
\rowcolor{black!5}
\textbf{Count} & An integer accumulated over time & \# of passes in Basketball; \# of steps in Cube \\
\addlinespace
\textbf{Location} & A position in a discrete / continuous space & Position of ball in Shell game / Tilt box  \\
\addlinespace
\rowcolor{black!5}
\textbf{Attribute} & A categorical or vector-valued property & Characters in Morse code; longest Latte art \\
\addlinespace
\midrule
\multicolumn{3}{l}{\emph{Structure}} \\
\addlinespace
\rowcolor{black!5}
\textbf{Atomic} & A single value at all time points & \# of pages in Book; \# of ingredients in Cooking\\
\addlinespace
\textbf{Sequence} & An ordered series of values over time & Typing in Keyboard; scoring order in Tennis \\
\addlinespace
\rowcolor{black!5}
\textbf{Set} & A subset of values, unordered & Distinct players in Volleyball; unpressed button in Numberpad \\
\addlinespace
\textbf{Dict} & A map binding each entity to a value & hits per player in Tennis; max shots made in Basketball \\
\bottomrule
\end{tabular}
\label{tab:state_types}
\end{table}
\begin{table}[ht!]
\centering\small
\caption{\textbf{Taxonomy of perceptual challenges }across diverse tasks (Table~\ref{tab:simulated_videos} and~\ref{tab:real_videos}).}
\begin{tabular}{l p{6.5cm} p{4cm}}
\toprule
\textbf{Challenge} & \textbf{Definition} & \textbf{Examples} \\
\midrule
\textbf{Occlusion} & The target is physically hidden behind other objects & Shell game; Cup stacking \\
\addlinespace
\rowcolor{black!5}
\textbf{Camera motion} & Viewpoint shifts disrupt the spatial reference frame & Basketball; Carousel\\
\addlinespace
\textbf{Homogeneity} & Multiple targets share identical appearance, making them hard to track individually & Cube; Lego \\
\addlinespace
\rowcolor{black!5}
\textbf{Symbolic decoding} & A continuous visual pattern must be segmented and mapped to discrete symbols & Keyboard; Graffiti \\
\addlinespace
\textbf{Multi-entity attribution} & Multiple objects act simultaneously, requiring state changes to be attributed to the correct one & Volleyball; NeuroTracker \\
\addlinespace
\rowcolor{black!5}
\textbf{Event ambiguity} & Visually similar events produce different state outcomes & Tightening bolts; Numberpad \\
\bottomrule
\end{tabular}
\label{tab:challenges}
\end{table}

\clearpage

\begin{figure*}[ht!]
\captionsetup[subfigure]{justification=centering}
\begin{subfigure}[t]{.33\textwidth}
\centering
\includegraphics[width=\textwidth]{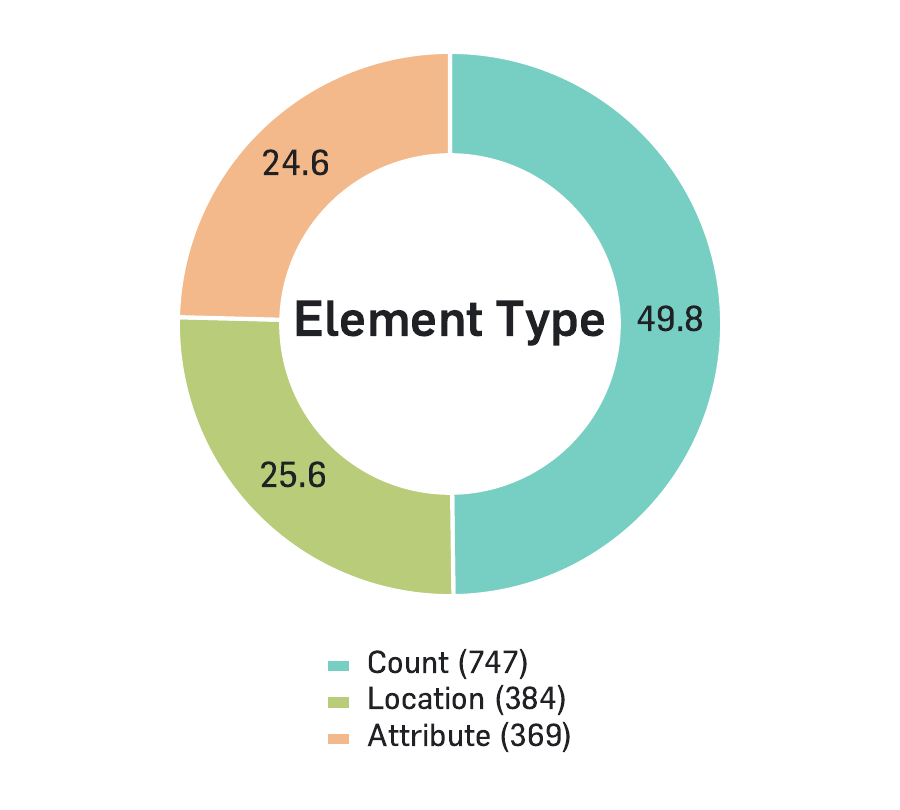}
\caption{Element type} 
\label{fig:stat:elem}
\end{subfigure}
\begin{subfigure}[t]{.33\textwidth}
\centering
\includegraphics[width=\textwidth]{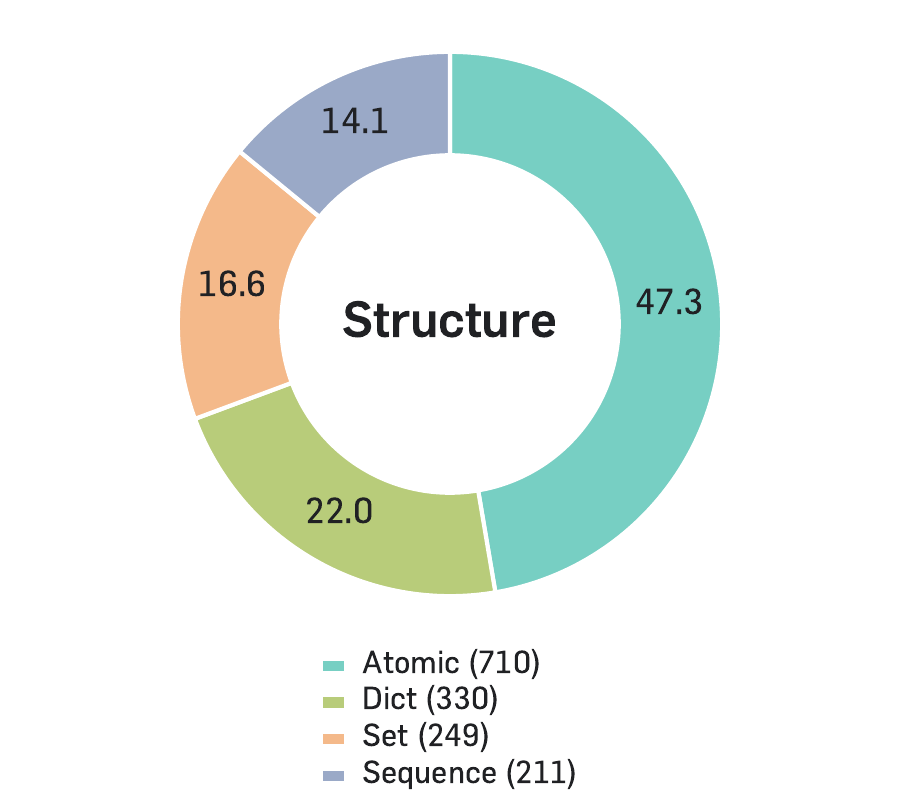}
\caption{Structure}
\label{fig:stat:strcture}
\end{subfigure}
\begin{subfigure}[t]{.33\textwidth}
\centering
\includegraphics[width=\textwidth]{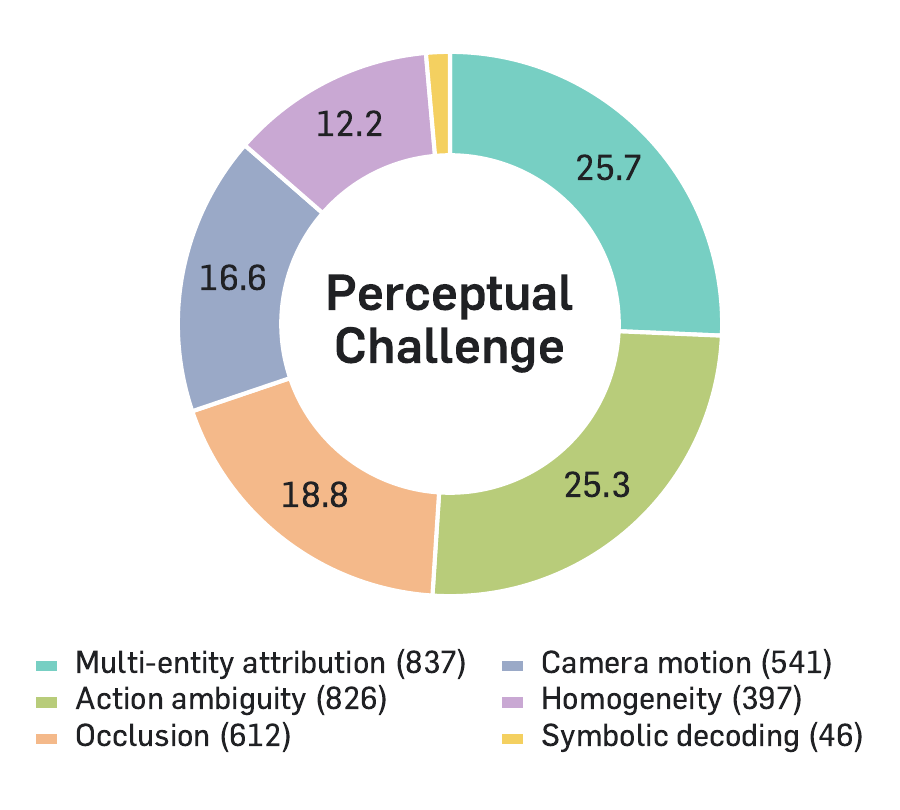}
\caption{Perceptual challenge} 
\label{fig:stat:perception}
\end{subfigure}
\begin{subfigure}{0.56\linewidth}
  \includegraphics[width=\linewidth]{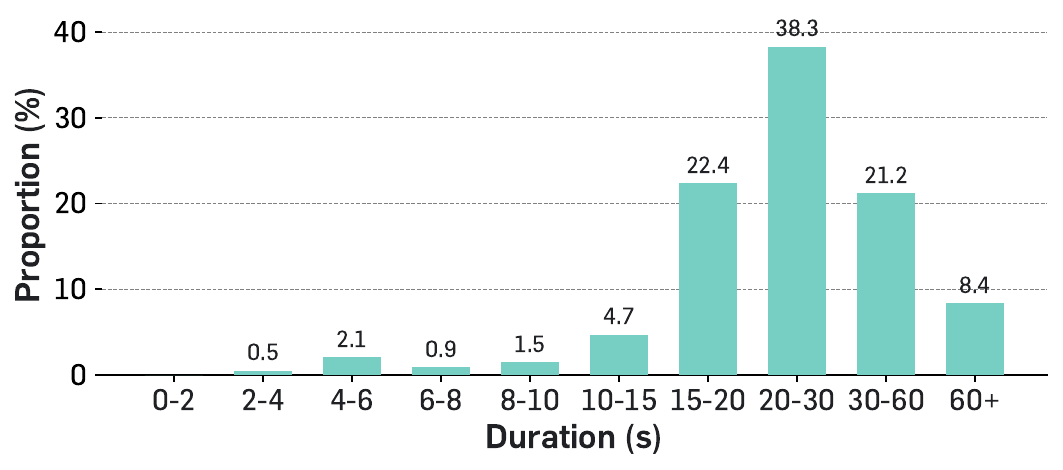}
  \caption{Duration}
  \label{fig:stat:duration}
\end{subfigure}~~
\begin{subfigure}{0.42\linewidth}
  \includegraphics[width=\linewidth]{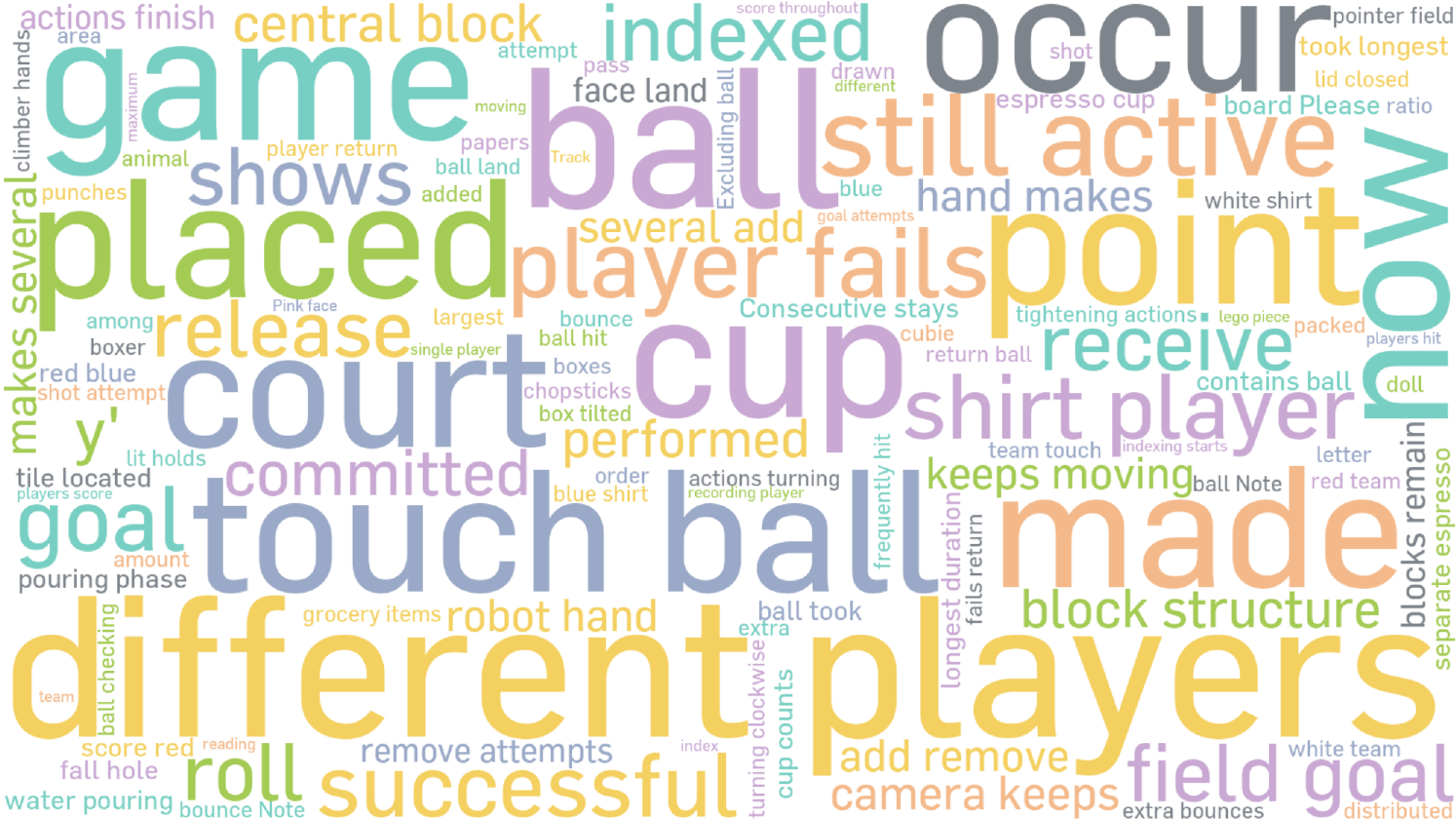}
  \caption{Question keywords}
\label{fig:stat:keywords}
\end{subfigure}
\caption{\textbf{Benchmark statistics of \sname.} 
We show the distribution of (a) element types, (b) state structures, (c) perceptual challenges, and (d) video durations, along with (e) a word cloud of question keywords. 
The benchmark exhibits a balanced distribution across all dimensions.}
\vspace{-0.2in}
\label{fig:stat}
\end{figure*}

\vspace{0.02in}\noindent
\textbf{Statistics.}
We provide the statistics of \sname in Figure~\ref{fig:stat}. As shown in the Figure~\ref{fig:stat:elem} and \ref{fig:stat:strcture}, our benchmark contains a balanced distribution across both element type and state structure, without skewing toward the simplest question types such as atomic count. Also, as shown in Figure~\ref{fig:stat:perception}, \sname includes various perceptual challenges (\eg, action ambiguity and camera motion). Moreover, as shown in Figure~\ref{fig:stat:duration}, a majority of our videos have a duration shorter than 30 seconds, which is much shorter than the context length of frontier models like Gemini-3.1 Pro or Gemini-3.0-Flash. Lastly, as shown in Figure~\ref{fig:stat:keywords}, our questions consist of diverse keywords, covering a wide range of situations. 

\clearpage

\subsection{Categories and example visualization}
\label{appen:subsec:examples}
\begin{figure*}[t]
\centering\small
\includegraphics[width=\textwidth]{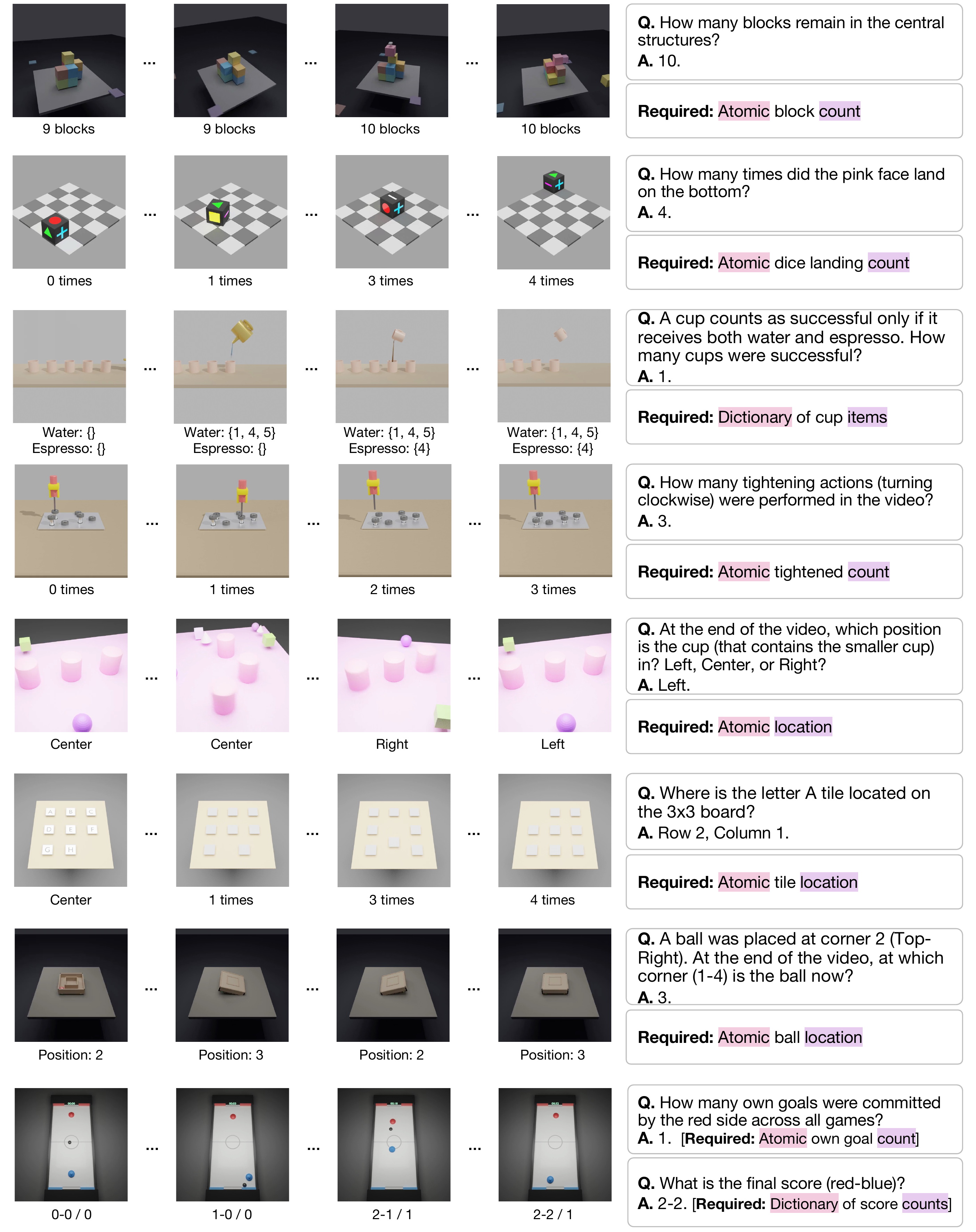}
\caption{
\textbf{Additional task examples in \sname synthesized with Blender.}
Each task requires different state complexity and involves diverse perceptual challenges.
}
\label{fig:more_qual_blender}
\end{figure*}

\begin{figure*}[t]
\centering\small
\includegraphics[width=\textwidth]{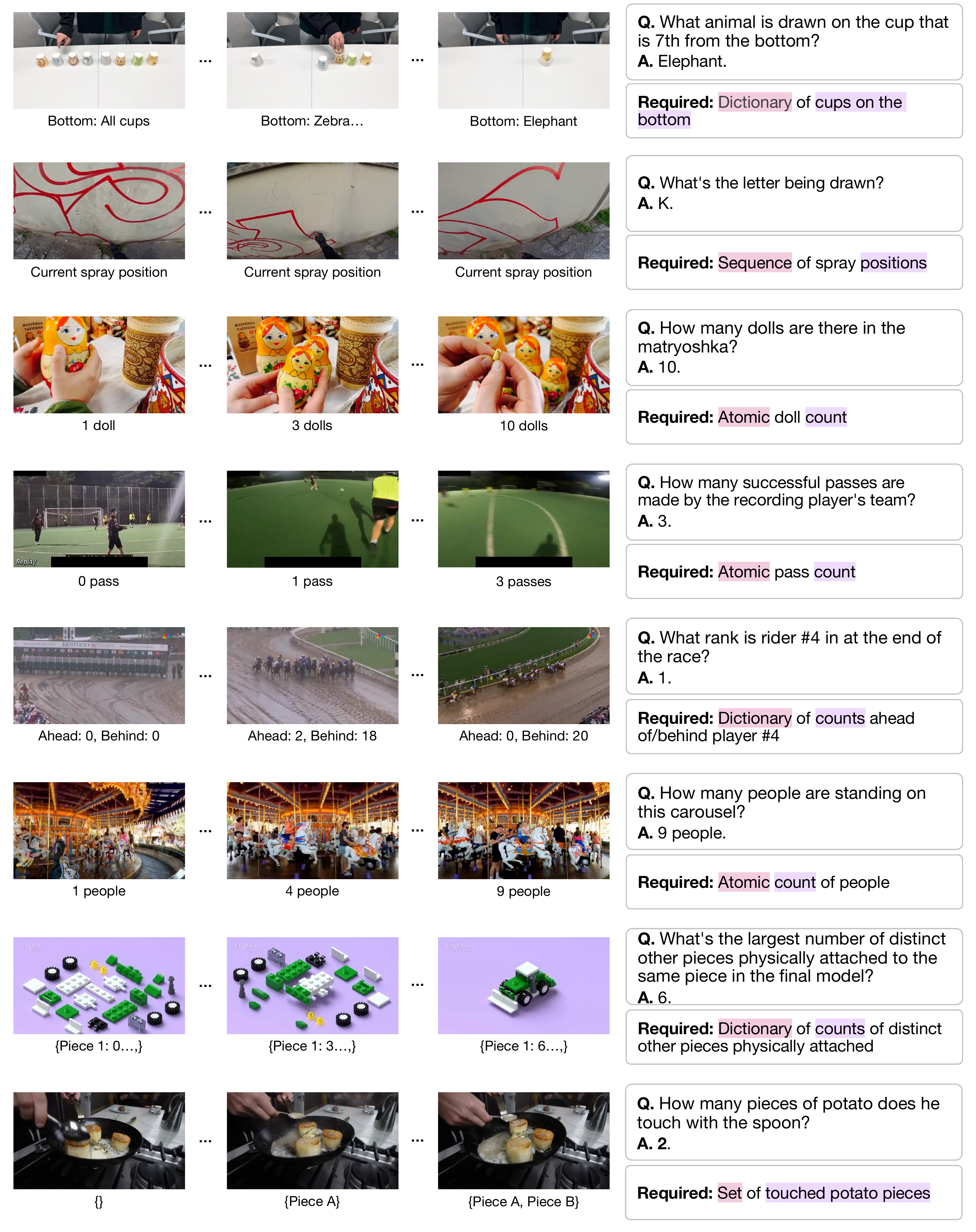}
\caption{
\textbf{Additional real-world task examples in \sname.} Each task requires different state complexity and has diverse perceptual challenges.
}
\label{fig:more_qual_real}
\end{figure*}

\textbf{Video categories and examples.}
Table~\ref{tab:simulated_videos} and~\ref{tab:real_videos} list the tasks and their descriptions: the former covers tasks implemented in Blender, while the latter covers tasks we recorded ourselves or curated from YouTube. 
We also visualize video examples for each category in 
Figure~\ref{fig:more_qual_blender} and \ref{fig:more_qual_real}.


\begin{table}[ht!]
\centering\small
\caption{\textbf{Simulated video tasks} rendered in Blender. \#Clips denotes the number of clips per task.}
\begin{tabular}{@{}p{2.5cm}p{10cm}c@{}}
\toprule
\textbf{Task} & \textbf{Description} & \textbf{\#Clips} \\
\midrule
Block count & Blocks in a 3D stack are shown in the video and blocks are randomly removed or added. The task is to predict the total number of blocks. & 50 \\
\addlinespace\rowcolor{black!5}
Rolling die & A die rolls across a surface, changing which face is up at each step. The task is to predict the total number of times a specific face is down. & 50 \\
\addlinespace
Americano making & There is first a water-pouring phase and then a separate espresso-pouring phase. A cup counts as successful only if it receives both water in the first phase and espresso in the second phase. The task is to infer the number of successful cups. & 50 \\
\addlinespace\rowcolor{black!5}
Tightening bolts & Bolts are randomly tightened or loosened throughout the video. The task is to predict the total number of tightening actions. & 50 \\
\addlinespace
Rotating shell game &  A ball is hidden under one of several cups, the cups are shuffled, and the task is to track and predict which cup the ball ends up under. The camera also rotates throughout the video. & 50 \\
\addlinespace\rowcolor{black!5}
Sliding puzzle & Like the 15-puzzle, tiles on a grid are slid into the empty space and each tile randomly moves throughout the video. The task is to predict the final position of a specific block. & 50 \\
\addlinespace
Tilt box & A box containing an object (e.g., a ball) is tilted in various directions, and the task is to predict where the object will end up. & 50 \\
\addlinespace\rowcolor{black!5}
Air hockey & A sequence of multiple air hockey plays. The task is to predict the total score, the longest game, or the number of own goals. & 50 \\
\addlinespace
Funnel drop & Multiple balls are released into funnels at different times, where they roll around before falling through the hole. Balls are indexed left-to-right in the last frame before any release from \texttt{ball\_1} through \texttt{ball\_6}. The task is to predict which ball took the longest time to fall through the hole after its own release. & 50 \\
\bottomrule
\end{tabular}
\label{tab:simulated_videos}
\end{table}

\begin{small}
\begin{longtable}{@{}p{2.5cm}p{9cm}c@{}}
\caption{\textbf{Real-world video tasks.} \#Clips denotes the number of unique video clips per task.}
\label{tab:real_videos} \\
\toprule
\textbf{Task} & \textbf{Description} & \textbf{\#Clips} \\
\midrule
\endfirsthead

\multicolumn{3}{c}{\textit{Table \thetable\ -- continued from previous page}} \\
\toprule
\textbf{Task} & \textbf{Description} & \textbf{\#Clips} \\
\midrule
\endhead

\midrule
\multicolumn{3}{r}{\textit{Continued on next page}} \\
\endfoot

\bottomrule
\endlastfoot

\rowcolor{black!5}
Book & A reader turns pages of a book either forward or backward. The task is to predict the net number of pages turned (signed). & 10 \\
\addlinespace
Tilt box & A real ball inside a box is tilted in various directions starting from a known corner. The task is to predict the corner where the ball ends up. & 10 \\
\addlinespace\rowcolor{black!5}
Shell game & Several cups are shuffled with a smaller cup hidden under one of them. The task is to predict the final position of the cup containing the smaller cup. & 10 \\
\addlinespace
Keyboard & A word is typed on a physical keyboard. The task is to identify the typed word. & 10 \\
\addlinespace\rowcolor{black!5}
Morse code & A light flashes a sequence in Morse code. The task is to decode the transmitted text. & 10 \\
\addlinespace
Numberpad & A sequence of numbers is pressed on a number pad. The task is to identify which two digits were not pressed. & 10 \\
\addlinespace\rowcolor{black!5}
Cup stacking & Cups with animal drawings are stacked in a tower. The task is to identify the animal on the cup at a given position from the bottom. & 10 \\
\addlinespace
Distributing items & Colored papers and chopsticks are distributed into cups. The task is to predict how many more items of a given type are needed for equal distribution. & 10 \\
\addlinespace\rowcolor{black!5}
Basketball & Real basketball plays including shots, passes, and 3-pointers. The task is to predict shot counts, field goal percentages, and per-player or per-team statistics. & 30 \\
\addlinespace
Bouldering & A climber moves on a wall with specific lit holds. The task is to count the total or maximum number of times the climber's hands or feet touch the lit holds. & 10 \\
\addlinespace\rowcolor{black!5}
Boxing & Two boxers exchange punches in a match. The task is to count punches by player, hand (left/right), or to determine the punch sequence. & 19 \\
\addlinespace
Carousel & A carousel ride filmed from a rider's or external viewpoint. The task is to count people, complete rounds, or exit passes. & 9 \\
\addlinespace\rowcolor{black!5}
Cooking \& barista & A chef or barista prepares foods and beverages such as sandwiches, burgers, noodles, espresso, latte, latte art, and sliced street food (yokan). The task is to count ingredients, cuts, cups, pours, or slices, and to identify preparation sequences or compare preparation times. & 23 \\
\addlinespace
Cube & A Rubik's cube is manipulated through several moves. The task is to track where a specific colored cubie ends up. & 16 \\
\addlinespace\rowcolor{black!5}
Eating contest & Contestants eat burgers in a competition. The task is to count consumed burgers or determine the finishing order. & 4 \\
\addlinespace
Graffiti & A person draws letters, words, or shapes on a wall. The task is to identify the drawn character or count the drawn shapes. & 16 \\
\addlinespace\rowcolor{black!5}
Horse racing & A horse race with multiple riders. The task is to predict final ranks of specific riders or count overtakes. & 4 \\
\addlinespace
Lego & A person assembles a Lego model. The task is to count specific colored pieces, connections, or evaluate symmetry of the final build. & 13 \\
\addlinespace\rowcolor{black!5}
Marching band & A marching band performs on a field. The task is to count players of a specific instrument crossing the centerline. & 4 \\
\addlinespace
Matryoshka & A set of nested Russian dolls is opened sequentially. The task is to count dolls or analyze headscarf and decoration patterns. & 8 \\
\addlinespace\rowcolor{black!5}
Order packing & Grocery items are packed into boxes or bags. The task is to count items, identify packing order, or reason about the minimum items to remove for visibility. & 21 \\
\addlinespace
Soccer & A real soccer game with multiple players. The task is to count goals, passes, possessions, or compute success rates. & 20 \\
\addlinespace\rowcolor{black!5}
Tennis & Real tennis matches with players exchanging shots. The task is to count returns, identify ball landing zones, or determine scoring order. & 20 \\
\addlinespace
Table tennis & Real table tennis matches between players. The task is to count hits, identify the server or winner, or track ball-table contacts. & 30 \\
\addlinespace\rowcolor{black!5}
Volleyball & A volleyball game with two teams. The task is to count total hits, distinct players touching the ball, or identify the team-contact sequence. & 20 \\
\addlinespace
Sokoban & A Sokoban puzzle is played with boxes pushed onto target destinations. The task is to count pushes, identify which box is pushed, or determine the remaining optimal moves. & 4 \\
\addlinespace\rowcolor{black!5}
NeuroTracker & A subset of moving balls is highlighted at the start. The task is to track and identify the originally highlighted balls at the end among numbered candidates. & 3 \\
\addlinespace
Memory card & A memory matching card game with face-down cards. The task is to identify matching pairs based on previously revealed cards. & 9 \\
\addlinespace
Guess Who & Multiple players kick or throw balls; the task is to identify who successfully lands the ball in the basket. & 21 \\
\end{longtable}
\end{small}

\clearpage
\clearpage

\subsection{Curation and Filtering Process}
\label{appen:subsec:curation}

 \textbf{Collecting and preprocessing strategies.}
For Blender videos, we set the video duration to 20 seconds for all tasks. For our analyses, we also synthesize shorter (5 seconds and 10 seconds) videos, but they are not included in our main benchmark and only used for the studies. For YouTube videos, we curate long-form footage and preprocess each video into clips with durations between 10 seconds and 1 minute, ensuring that no clip contains ambiguous events caused by clip boundaries. For example, in soccer clips, each video clearly shows whether a shot resulted in a goal. For recorded videos, all videos featuring identifiable persons were recorded by the authors with explicit consent for research and public release. 

\vspace{0.02in}\noindent
\textbf{Question-answer generation with a human-in-the-loop process.}
For each video clip, we design various questions to ensure that each question requires a different minimum amount of information (\ie, state complexity) to answer. For example, our questions include keywords such as ``second-to-last'' or ``total'' count, requiring the model to track information over the entire video. For videos that contain interactions among multiple entities with identical appearances, we construct questions that include keywords such as ``how many people'' or ``who performed the action most'', as these require distinguishing each entity, which is possible only if the model keeps track of the trajectory of each entity over time.
Due to the lack of ground truth in video metadata for our hand-designed questions, as well as the limited visual state tracking capability of current MLLMs, automatic annotation for QA pairs is largely infeasible. We therefore manually labeled the answers to all questions. To ensure accuracy and eliminate ambiguity, every QA pair underwent at least two rounds of human validation. Any QA pair that human reviewers still deemed ambiguous after multiple rounds of review was removed from the final benchmark.

\vspace{0.02in}\noindent
\textbf{Multiple-choice question (MCQ) distractors.}
For MCQs, distractors are generated from plausible alternative states that could result from common tracking errors, rather than from semantically unrelated answers. Specifically, we provide the questions and answer choices without the video stream and check whether the model can predict the answer. In such cases, we reconstruct the other answer choices to avoid such shortcuts.

\vspace{0.02in}\noindent
\textbf{Labeling and filtering.}
To analyze performance with a breakdown, we label each question using our taxonomy. Each label is double-checked by a reviewer who has not labeled the question. We use the more formal definitions of each taxonomy in Table~\ref{tab:state_types} and \ref{tab:challenges} to remove any ambiguity in labeling.

\clearpage
\section{Evaluation Setup Details}
\label{appen:eval}
\begin{figure*}[h]
\centering
    \includegraphics[width=.85\textwidth]{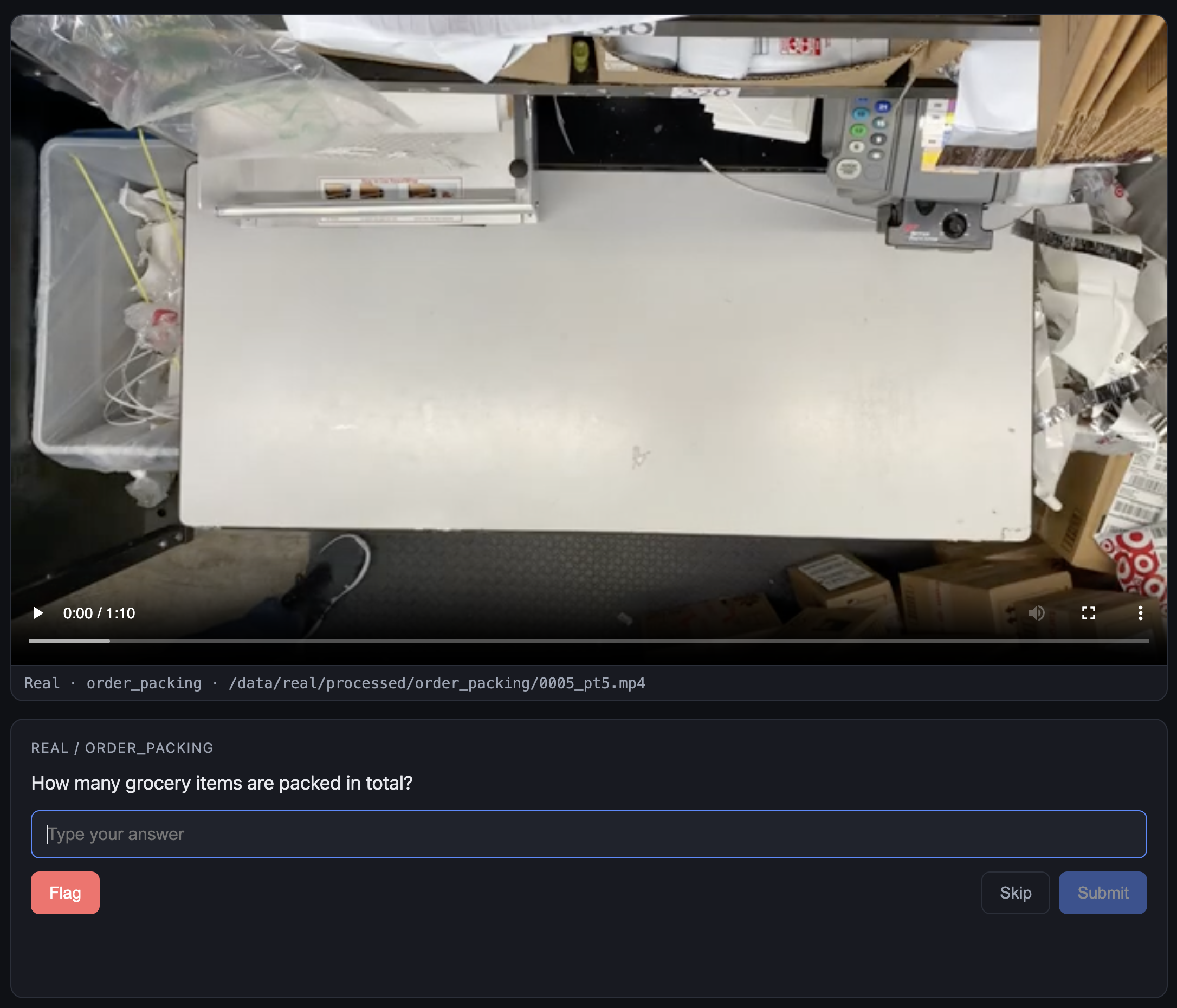}
    \caption{\textbf{Human evaluation UI.}}
    \label{fig:eval_ui}
\end{figure*}

\textbf{Human evaluation.}
To measure human performance, we internally built a website for evaluation. The evaluation was conducted by participants including the authors, but excluding those who had contributed specific videos or questions, to avoid any prior knowledge or information leakage. Participants were allowed to watch each video multiple times and think freely, but were strictly limited to a single answer per question. The ground-truth answer was never shown during the task, and each response was locked once submitted. We visualize our evaluation UI in Figure~\ref{fig:eval_ui}. 

\vspace{-0.1in}\noindent
\textbf{Chance-level performance.}
Following VSI-Bench \citep{yang2024think}, we consider two types of chance-level performance: (a) random and (b) frequency-based. For (a), we assume no access to the answer distribution and guess uniformly at random. We compute this accuracy only for multiple-choice questions (MCQs), not for numerical answers (NA). For (b), we estimate the empirical answer distribution $p$ over both MCQs and NAs, and report the expected score of the best deterministic predictor: always predicting the most frequent answer (mode) for accuracy, and the optimal constant for MRA.
\begin{align*}
\text{Acc}_{\text{rand}}^{\text{mcq}} &= \tfrac{1}{k}, \\
\text{Acc}_{\text{freq}}^{\text{mcq/num}} &= \max_i p_i, \\
\text{MRA}_{\text{freq}}^{\text{num}} &= \max_{c \in [\ell, h]} \mathbb{E}_a\!\left[\text{MRA}_{\text{thr}}(c, a)\right].
\end{align*}
Here MRA denotes the threshold-based MRA following OpenEQA \citep{majumdar2024openeqa}, with thresholds $\theta \ \in \{0.5, 0.55, \ldots, 0.95\}$. Note that we compute these accuracies separately for each question type, since the magnitude of the answers can vary substantially across them. For example, counting questions typically have maximum values below 10, whereas success/failure rates range from 0 to 100.

\clearpage
\section{Additional Results}

\subsection{Results across video sources}

We provide the performance decomposition of different models on \sname across the three video sources: Blender, Recorded, and YouTube.

\begin{table*}[h!]
    \centering
    \caption{\textbf{Evaluation on \sname{} by video category.} Scores report the reparsed MRA-with-MCQ metric, broken down by video source: \emph{YouTube} (in-the-wild clips), \emph{Synthetic} (rendered tasks), and \emph{Recorded} (lab-recorded tasks). \tabledarkhl{Dark gray} indicates the best result among all models and \tablelighthl{light gray} indicates the best result among open-sourced models. Ranks are computed separately within proprietary API models (1--4) and within open-sourced models (1--20, pooling Thinking and Instruct); baselines are not ranked.}
    \footnotesize
    \setlength\tabcolsep{3pt}%
    \renewcommand{\arraystretch}{1.15}
\begin{tabular}{r|cc|ccc}
 & & &
 \rotatebox{55}{YouTube} &
 \rotatebox{55}{Synthetic} &
 \rotatebox{55}{Recorded} \\
Methods & Rank & Avg. & \multicolumn{3}{c}{\cellcolor{stateelementtable}Video Category} \\
\hline
\rowcolor{tablesectionbg}
\multicolumn{1}{l|}{\textcolor{black}{\textit{Baselines}}} & & & & & \\
Chance Level (Random)         & - & 26.1 & 25.7 & 26.4  & 26.2  \\
Chance Level (Frequency)      & - & 37.8 & 38.2 & 37.7 & 34.3  \\
Human Performance             & - & \textbf{90.5} & 86.5 & 98.0 & 82.8 \\
\hline
\rowcolor{tablesectionbg}
\multicolumn{1}{l|}{\textcolor{black}{\textit{Proprietary Models (API)}}} & & & & & \\
Gemini-3.1 Pro (low) \citep{google2026gemini31pro}    & \cellcolor{oai-green-200}{1} & \cellcolor{oai-gray-500}{44.4} & \cellcolor{oai-gray-500}{42.6} & 38.5 & \cellcolor{oai-gray-500}{54.1} \\
Gemini-3.1 Pro (high) \citep{google2026gemini31pro}   & \cellcolor{oai-green-100}{2} & 43.9 & 42.1 & \cellcolor{oai-gray-500}{41.6} & 49.9 \\
Gemini-3.0 Flash (low) \citep{google2025gemini3flash} & 3 & 39.8 & 33.4 & 40.3 & 52.2 \\
Gemini-3.0 Flash (high) \citep{google2025gemini3flash} & 4 & 38.8 & 33.2 & 36.6 & 52.5 \\
\hline
\rowcolor{tablesectionbg}
\multicolumn{1}{l|}{\textcolor{black}{\textit{Open-sourced Models Thinking}}} & & & & & \\
MiMo-VL-7B \citep{Yue2025MiMoVLTR}                   & 11 & 31.2 & 35.3 & 24.3 & 34.3 \\
InternVL3.5-8B-Thinking \citep{Wang2025InternVL35AO} & 13 & 30.2 & 29.5 & 30.4 & 35.5 \\
GLM-4.1V-9B-Thinking \citep{zhipu2025glm41v}         & 14 & 30.2 & 31.8 & 26.4 & \cellcolor{oai-gray-300}{37.4} \\
Qwen3VL-8B-Thinking \citep{Bai2025Qwen3VLTR}         & 18 & 28.2 & 29.3 & 26.1 & 29.8 \\
Qwen3VL-4B-Thinking \citep{Bai2025Qwen3VLTR}         & 19 & 26.0 & 26.7 & 23.7 & 33.3 \\
\hline
\rowcolor{tablesectionbg}
\multicolumn{1}{l|}{\textcolor{black}{\textit{Open-sourced Models Instruct}}} & & & & & \\
LLaVA-OV-2-8B (frames) \citep{llavaonevision2} & \cellcolor{oai-green-200}{1} & \cellcolor{oai-gray-300}{35.1} & \cellcolor{oai-gray-300}{40.6} & 27.7 & 29.0 \\
LLaVA-OV-2-8B (codec) \citep{llavaonevision2}  & \cellcolor{oai-green-100}{2} & 35.0 & 40.5 & 27.1 & 32.0 \\
Molmo2-4B \citep{molmo2}                       & 3 & 34.4 & 32.4 & 37.1 & 34.7 \\
Cambrian-S-7B \citep{yang2026cambrians}        & 4 & 34.2 & 32.5 & \cellcolor{oai-gray-300}{39.6} & 18.7 \\
Molmo2-8B \citep{molmo2}                       & 5 & 34.0 & 35.5 & 31.5 & 34.0 \\
Qwen3VL-8B \citep{Bai2025Qwen3VLTR}            & 6 & 33.2 & 36.9 & 29.2 & 23.9 \\
InternVL3.5-2B \citep{Wang2025InternVL35AO}    & 7 & 31.8 & 31.7 & 33.1 & 26.0 \\
Cambrian-S-3B \citep{yang2026cambrians}        & 8 & 31.8 & 33.2 & 30.0 & 29.4 \\
VITA-1.5-7B \citep{fu2025vita15}               & 9 & 31.5 & 34.1 & 28.6 & 25.0 \\
Qwen3VL-4B \citep{Bai2025Qwen3VLTR}            & 10 & 31.3 & 34.1 & 27.1 & 30.5 \\
InternVL3.5-8B \citep{Wang2025InternVL35AO}    & 12 & 30.6 & 32.7 & 27.6 & 30.1 \\
Qwen3VL-2B \citep{Bai2025Qwen3VLTR}            & 15 & 29.4 & 28.8 & 31.7 & 21.5 \\
Cambrian-S-1.5B \citep{yang2026cambrians}      & 16 & 29.3 & 31.6 & 26.5 & 25.7 \\
LLaVA-OV-7B \citep{li2024llava}                & 17 & 28.6 & 27.5 & 30.7 & 26.8 \\
LLaVA-OV-0.5B \citep{li2024llava}              & 20 & 21.3 & 16.6 & 27.8 & 25.0 \\
\end{tabular}
\label{tab:main_category}
\vspace{0.03in}
\begin{minipage}{0.98\textwidth}
\scriptsize\emph{Note.}
Open-sourced rows use the same best frame setting as Table~\ref{tab:main}; the corrected \texttt{guess\_who\_make\_it} subset is assigned to the YouTube source bucket.
\end{minipage}
\end{table*}

\clearpage

\subsection{Text transcription examples}
\label{appen:transcription}

\begin{figure}[ht!]
    \centering
\begin{tcolorbox}
  [colback=softpeach!5,
  colframe=softpeach!70!black!85,
  colbacktitle=softpeach!75!black!85,
  coltitle=white,
  boxrule=0.7pt,
  arc=2mm,
  title=\small{\textbf{Q.} During the roll, how many times did the Pink face land on the bottom? (Consecutive stays count as 1, including the initial position)}, fontupper=\footnotesize, breakable]
    \includegraphics[width=\textwidth]{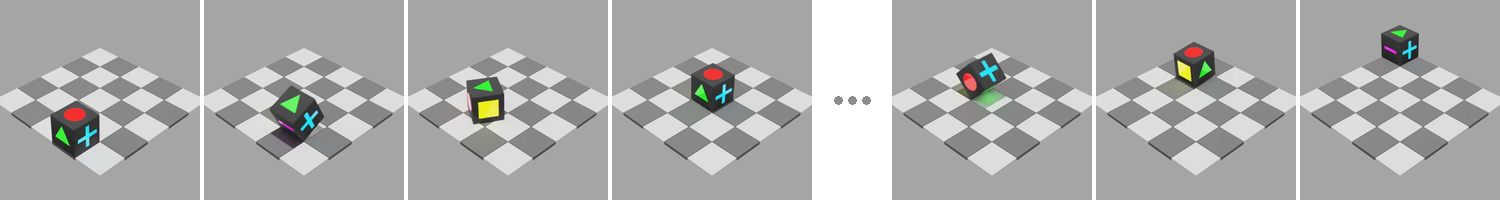}\\

\textbf{[Text description.]}

The die starts with the following three visible faces: Red on top, Green in front, and Blue on the right.
The die is then moved as follows: roll up, roll down, roll left, …
After each move, the three visible faces become: (Green, Pink, Blue), (Red, Green, Blue), (Blue, Green, Pink) ...\\

\textbf{[Gemini-3.1 Pro Thinking Trace]}

To determine how many times the Pink face touched the floor, we first need to deduce the opposite pairs of faces on the die.
\begin{enumerate}[leftmargin=*,itemsep=0mm]
\item \textbf{Initial State:} The visible faces are Top = Red, Front = Green, Right = Blue.
\item \textbf{Move 1 (roll up):} The visible faces become Top = Green, Front = Pink, Right = Blue. When rolling up, the old Bottom face becomes the new Front face. Since the new Front is Pink, the initial Bottom face
was Pink. Because the initial Top face was Red, we can conclude that \textbf{Red and Pink are opposite faces}.
\item \textbf{Move 3 (roll left):} From Move 2 (Top = Red, Front = Green, Right = Blue), rolling left makes the old Right (Blue)
the new Top, and the old Bottom (Pink) the new Right. The visible faces are Top = Blue, Front = Green, Right = Pink.
\item \textbf{Move 4 (roll up):} Rolling up from Move 3 makes the old Bottom the new Front. The visible faces are Top = Green,
Front = Yellow, Right = Pink. Since the new Front is Yellow, the old Bottom was Yellow. Because the Top in Move 3
was Blue, \textbf{Blue and Yellow are opposite faces}.
\item By elimination, the remaining two faces, \textbf{Green and White, are opposite faces}.
Since Red and Pink are opposite faces, the Pink face is on the bottom (touching the floor) exactly when the Red face is on the top. We can track the Top face at the initial position and after each of the 19 moves by looking at the first color in each of
the provided visible face triplets:
    \begin{itemize}[leftmargin=*,itemsep=0mm]
    \item \textbf{Initial}: \textbf{Red} (Pink is on the bottom)
    \item Move 1: Green - Move 2: \textbf{Red} (Pink is on the bottom) 
    \item Move 3: Blue
    \item Move 4: Green 
    \item Move 5: Green 
    \item Move 6: \textbf{Red} (Pink is on the bottom)
    \item Move 7: \textbf{Red} (Pink is on the bottom) 
    \item Move 8: \textbf{Red} (Pink is on the bottom)
    \item ...
    \end{itemize}
\end{enumerate}
Now, we count the distinct times the Pink face was on the bottom, treating consecutive stays as 1:
\begin{enumerate}[leftmargin=*,itemsep=0mm]
\item Initial position
\item Move 2
\item Moves 6, 7, and 8 (counted as 1 distinct time since they are consecutive)
\item Move 17
\end{enumerate}
The Pink face touched the floor 4 distinct times, including the initial position.
\textbf{Answer: 4}~\GOOD \quad \textcolor{ggreen}{\textbf{GT: 4}}
\end{tcolorbox}
    \caption{\textbf{Text transcription and thinking trace summary} for the rolling die task.}
    \label{fig:text_transcription_dice}
\end{figure}

\begin{figure}[ht!]
    \centering
    \begin{tcolorbox}[
      colback=softpeach!5,
      colframe=softpeach!70!black!85,
      colbacktitle=softpeach!75!black!85,
      coltitle=white,
      boxrule=0.7pt,
      arc=2mm,
      title={\small\textbf{Q.} At the end of the video, which position is the Cup B (that contains the ball) in?},
      fontupper=\footnotesize
    ]
        \includegraphics[width=\textwidth]{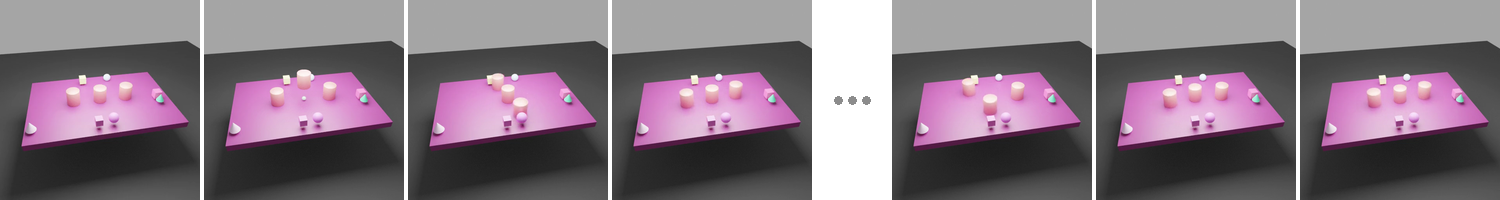} \\
        
        \textbf{[Text description.]}
        
        Three cups are placed in order — Left, Middle, Right — and the ball starts under the Middle cup.
        Swaps (each swap exchanges the positions of the two listed cups): (Left, Middle), (Left, Middle), (Left, Middle), (Left, Right)... \\

        \textbf{[Gemini-3.1 Pro Thinking Trace]}
        
        To determine the final position of the ball, we can track its location after each swap.

        \begin{enumerate}[leftmargin=*,itemsep=0mm]
        \item \textbf{(Left, Middle)} - The ball moves from Middle to \textbf{Left}.
        \item \textbf{(Left, Middle)} - The ball moves from \textbf{Left} to \textbf{Middle}.
        \item \textbf{(Left, Middle)} - The ball moves from Middle to \textbf{Left}.
        \item \textbf{(Left, Right)} - The ball moves from Left to \textbf{Right}.
        \item \textbf{(Middle, Right)} - The ball moves from Right to \textbf{Middle}.
        \item \textbf{(Middle, Right)} - The ball moves from Middle to \textbf{Right}.
        \item \textbf{(Middle, Right)} - The ball moves from Right to \textbf{Middle}.
        \item \textbf{(Left, Right)} - The ball is not involved in this swap, so it stays at \textbf{Middle}.
        \item ...
        \item \textbf{(Left, Right)} - The ball is not involved in this swap, so it stays at \textbf{Middle}.
        \end{enumerate}
    \textbf{Answer: Center}~\GOOD \quad  \textcolor{ggreen}{\textbf{GT: Center}}
    \end{tcolorbox}
    \caption{\textbf{Text transcription and thinking trace summary} for the shell game task.}
    \label{fig:text_transcription_shell}
    \vspace{-0.1in}
\end{figure}

\begin{figure}[ht!]
    \centering
    \begin{tcolorbox}[
      colback=softpeach!5,
      colframe=softpeach!70!black!85,
      colbacktitle=softpeach!75!black!85,
      coltitle=white,
      boxrule=0.7pt,
      arc=2mm,
      title={\small\textbf{Q.}Which corner will the ball end up at?},
      fontupper=\footnotesize
    ]
        \includegraphics[width=\textwidth]{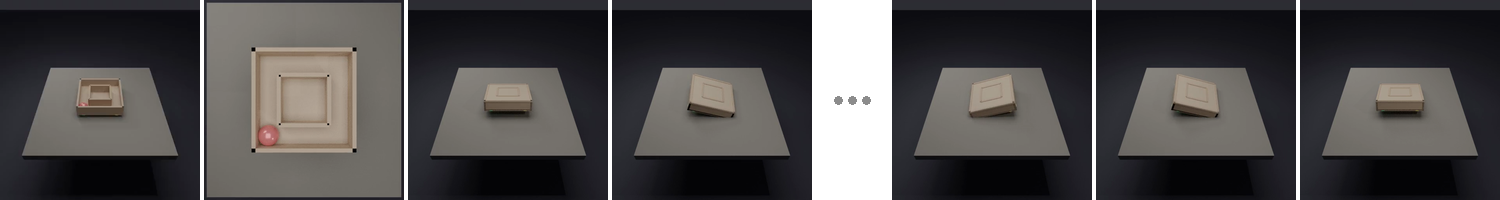} \\
        
        \textbf{[Text description.]}
        
        Label the corners of a box as follows:
        
        1 --- 2 
        
        |~~~~~~~~|
        
        4 --- 3
        
        Place a ball at corner 2. Then tilt the box in the following directions in order: Down, Left, Down, Up, Right, Down. \\

        \textbf{[Gemini-3.1 Pro Thinking Trace]}
        
        To determine the final position of the ball, let's map out the corners of the box based on the provided diagram:
        Corner 1: Top-Left, Corner 2: Top-Right, Corner 3: Bottom-Right, Corner 4: Bottom-Left
        
        The ball starts at \textbf{corner 2} (Top-Right). Now we follow the sequence of tilts:

        \begin{enumerate}[leftmargin=*,itemsep=0mm]
        \item \textbf{Down}: The ball rolls from the top edge to the bottom edge. From Top-Right (2), it goes to Bottom-Right (\textbf{3}).
        \item \textbf{Left}: The ball rolls from the right edge to the left edge. From Bottom-Right (3), it goes to Bottom-Left (\textbf{4}).
        \item \textbf{Down}: The ball is already on the bottom edge, so it stays at Bottom-Left (\textbf{4}).
        \item \textbf{Up}: The ball rolls from the bottom edge to the top edge. From Bottom-Left (4), it goes to Top-Left (\textbf{1}).
        \item \textbf{Right}: The ball rolls from the left edge to the right edge. From Top-Left (1), it goes to Top-Right (\textbf{2}).
        \item \textbf{Down}: The ball rolls from the top edge to the bottom edge. From Top-Right (2), it goes to Bottom-Right (\textbf{3}).
        \end{enumerate}
    \textbf{Answer: 3}~\GOOD \quad \textcolor{ggreen}{\textbf{GT: 3}}
    \end{tcolorbox}
    \caption{\textbf{Text transcription and thinking trace summary} for the tilt box task.}
    \label{fig:text_transcription_tilt}
\end{figure}

We provide the full text transcription results of three Blender tasks (rolling die, shell game, and tilt box), along with their reasoning traces from Gemini-3.1 Pro \citep{google2026gemini31pro} in Figure~\ref{fig:text_transcription_dice}, \ref{fig:text_transcription_shell}, and \ref{fig:text_transcription_tilt}. As shown in the figure, the model can do tracking and reasoning near perfectly.

\clearpage
\subsection{Additional Failure cases}
\label{appen:failure}

\textbf{Quantitative analysis details.}
From each video category, we select multiple questions, each requiring a different state element and structure, to ensure that our analysis covers the full diversity of question types in our benchmark. This yields a total of 70 questions, providing a comprehensive basis for analyzing failure cases across video content, state elements, and structures.

\textbf{Illustration of additional failure cases.}
In what follows, we illustrate the failure cases from Gemini-3.1 Pro with their thinking traces.

\begin{figure}[ht!]
    \centering
    \begin{tcolorbox}[
      colback=softpeach!5,
      colframe=softpeach!70!black!85,
      colbacktitle=softpeach!75!black!85,
      coltitle=white,
      boxrule=0.7pt,
      arc=2mm,
      title={\small\textbf{Q.} What is the final score (red-blue)? (A) 3-1 (B) 2-2 (C) 4-0 (D) 0-4},
      fontupper=\footnotesize
    ]
        \includegraphics[width=\textwidth]{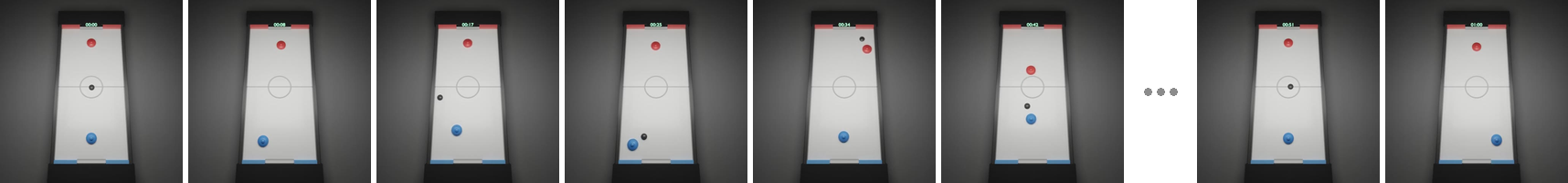} \\
        \textbf{A.} Based on the video, here is the breakdown of the goals:
        \begin{enumerate}[leftmargin=*,itemsep=0mm]
            \item At 00:08, the puck goes into the bottom goal. Red scores. (Red 1 - Blue 0)
            \item At 00:18, \percpcode{the puck goes into the top goal}. \elemcode{Blue scores. (Red 1 - Blue 1)}
            \item At 00:27, \percpcode{the puck goes into the top goal}. \elemcode{Blue scores. (Red 1 - Blue 2)}
            \item At 00:39, the puck goes into the bottom goal. Red scores. (Red 2 - Blue 2)
            \item At 00:49, the puck goes into the bottom goal. Red scores. (Red 3 - Blue 2)
            \item At 00:58, the puck goes into the bottom goal. Red scores. (Red 4 - Blue 2)
        \end{enumerate}
        The final score is Red 4, Blue 2. However, this option is not available. Let's re-examine the video carefully.
        It appears there might be a misunderstanding of the video's events or the provided options are incorrect. Given
        the strict instruction to only output a single letter from the choices, and none of the choices match the observed
        4-2 score, I will select the closest option or re-evaluate.

        Let's re-watch.
        
        00:08 - Red scores (1-0), 00:18 - \elemcode{Blue scores (1-1)}, 00:27 - \elemcode{Blue scores (1-2)}, 00:39 - Red scores (2-2), 00:49 - Red scores (3-2), 00:58 - Red scores (4-2)

        \textbf{Answer: B}~\BAD \quad \textcolor{ggreen}{\textbf{GT: C}}
    \end{tcolorbox}
    \caption{\textbf{Additional failure examples.} We highlight phrases and frames related to state extraction in \elemcode{purple} and failures in visual perception in \percpcode{green}. \textit{Failure reason:} Event recognition. The model misunderstands the ball reflected off the red-side wall as a goal.}
    \label{fig:addtional_failure_p1}
\end{figure}

\begin{figure}[ht!]
    \centering
    \begin{tcolorbox}[
      colback=softpeach!5,
      colframe=softpeach!70!black!85,
      colbacktitle=softpeach!75!black!85,
      coltitle=white,
      boxrule=0.7pt,
      arc=2mm,
      title={\small\textbf{Q}. Which two numbers on the number pad were not pressed? Answer with two numbers separated by a comma. (A) 5,9 (B) 2,9 (C) 1,9 (D) 0,9},
      fontupper=\footnotesize
    ]
        \includegraphics[width=\textwidth]{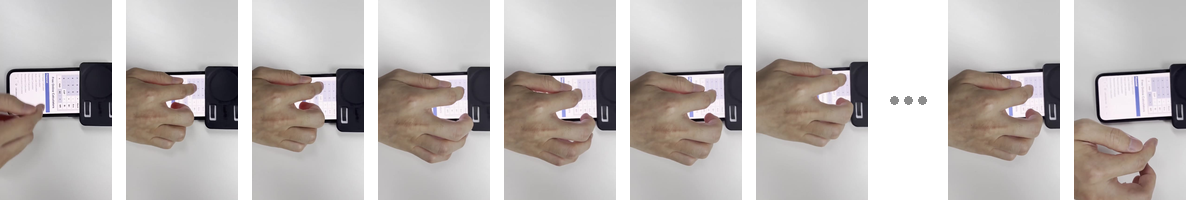} \\
        \textbf{A.} Based on a careful observation of the video, here is the sequence of actions on the numberpad:
        \begin{enumerate}[leftmargin=*,itemsep=0mm]
         \item The user distinctly \percpcode{taps the numbers 1, 2, 3} (frames 00:01 to 00:03).
         \item The user distinctly \percpcode{taps the numbers 4, 5, 6} (frames 00:04 to 00:06).
         \item The user distinctly taps the numbers \percpcode{7, 8} (frames 00:08 to 00:09).
         \item After \percpcode{pressing 8, the finger moves downwards to press the decimal point '.'} (frame 00:12). During this movement, the finger passes over the 0 key (frame 00:11) but \elemcode{does not perform a distinct tapping motion on it}; it appears to be a continuous sliding motion towards the decimal point.
         \item The finger never moves to or presses the 9 key.
        \end{enumerate}
        Therefore, the two numbers that were not pressed are 0 and 9. 

        \textbf{Answer: D}~\BAD \quad \textcolor{ggreen}{\textbf{GT: A}}
    \end{tcolorbox}
    \caption{\textbf{Additional failure examples.} We highlight phrases and frames related to state extraction in \elemcode{purple} and failures in visual perception in \percpcode{green}. \textit{Failure reason:} Event recognition. Occlusion makes the model infer an incorrect click action, leading to hallucinations.}
    \label{fig:addtional_failure_p2}
\end{figure}

\begin{figure}[ht!]
    \centering
    \begin{tcolorbox}[
      colback=softpeach!5,
      colframe=softpeach!70!black!85,
      colbacktitle=softpeach!75!black!85,
      coltitle=white,
      boxrule=0.7pt,
      arc=2mm,
      title={\small\textbf{Q.} Does the eighth doll wear a headscarf? (A) No (B) Yes},
      fontupper=\footnotesize
    ]
        \includegraphics[width=\textwidth]{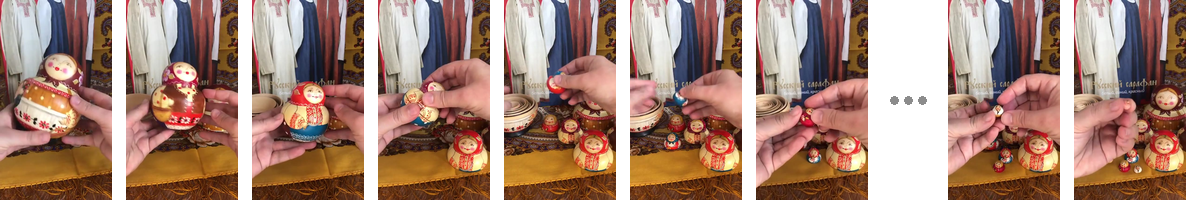} \\
        Based on the video, we can count the dolls as they are opened:
        \begin{enumerate}[leftmargin=*,itemsep=0mm]
        \item The first (outermost) doll is opened at 00:00.
        \item The second doll is revealed at 00:05.
        \item The third doll is revealed at 00:16.
        \item The fourth doll is revealed at 00:25.
        \item The fifth doll is revealed at 00:34.
        \item The sixth doll is revealed at 00:45.
        \item \elemcode{The seventh doll is revealed at 00:58}.
        \item \elemcode{The eighth doll is revealed at 01:07}.
    \end{enumerate}
    When the eighth doll is held up and shown clearly around 01:11, it has a white face, a blue body, and a red painted head covering, which represents a headscarf, just like the larger dolls.
    Therefore, the eighth doll does wear a headscarf. \textbf{Answer}: B~\BAD \quad \textcolor{ggreen}{\textbf{GT: A}}
    \end{tcolorbox}
    \caption{\textbf{Additional failure examples.} We highlight phrases and frames related to state extraction in \elemcode{purple} and failures in visual perception in \percpcode{green}. \textit{Failure reason:} Event recognition. The model misses the reveal of the seventh doll at 00:50, leading to an incorrect prediction.}
    \label{fig:addtional_failure_p3}
\end{figure}

\begin{figure}[ht!]
    \centering
    \begin{tcolorbox}[
      colback=softpeach!5,
      colframe=softpeach!70!black!85,
      colbacktitle=softpeach!75!black!85,
      coltitle=white,
      boxrule=0.7pt,
      arc=2mm,
      title={\small\textbf{Q}. At the end of the video, which position is the Cup B (that contains the ball) in? Left, Center, or Right? (A) Right (B) Center (C) Left},
      fontupper=\footnotesize
    ]
        \includegraphics[width=\textwidth]{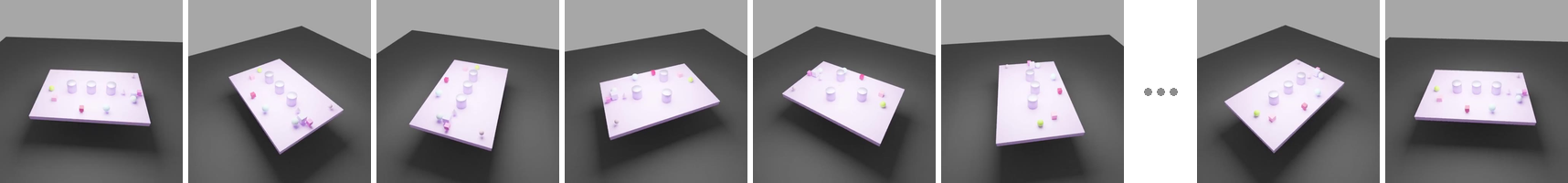} \\
        \textbf{A.} Based on a careful tracking of the cup containing the ball throughout the video:
        \begin{enumerate}[leftmargin=*,itemsep=0mm]
        \item 00:00: The ball is initially placed under the Center cup.
        \item 00:02 - 00:04: The Center cup \percpcode{moves to the Top} position.
        \item 00:05 - 00:08: The cup \percpcode{moves from the Top to the Right, and then to the Bottom-Right position}.
        \item 00:09 - 00:12: The cup \percpcode{moves from Bottom-Right to Bottom-Center, and then back to the Center position}.
        \item 00:13 - 00:16: The cup \percpcode{moves from the Center to the Right, and then to the Top-Right position}.
        \item 00:17 - 00:19: The cup \percpcode{moves from Top-Right to the Center position} as the cups realign.
        \end{enumerate}   
        The final position of the cup with the ball is in the Center.\\
        \textbf{Answer:} B~\BAD\quad\textcolor{ggreen}{\textbf{GT: C}}
    \end{tcolorbox}
    \caption{\textbf{Additional failure examples.} We highlight phrases and frames related to state extraction in \elemcode{purple} and failures in visual perception in \percpcode{green}. \textit{Failure reason:} Entity association. Continuous camera rotation changes relative positions (e.g., top, bottom, left, and right), leading to incorrect entity association.}
    \label{fig:addtional_failure_p4}
\end{figure}

\begin{figure}[ht!]
    \centering
    \begin{tcolorbox}[
      colback=softpeach!5,
      colframe=softpeach!70!black!85,
      colbacktitle=softpeach!75!black!85,
      coltitle=white,
      boxrule=0.7pt,
      arc=2mm,
      title={\small\textbf{Q.} At the first frame of the clip, index the six boxes by reading each row from left to right, then moving from the top row to the bottom row (indexing starts from 1). Which indexed box is pushed last before the level is completed? (A) box 5 (B) box 2 (C) box 4 (D) box 1},
      fontupper=\footnotesize
    ]
        \includegraphics[width=\textwidth]{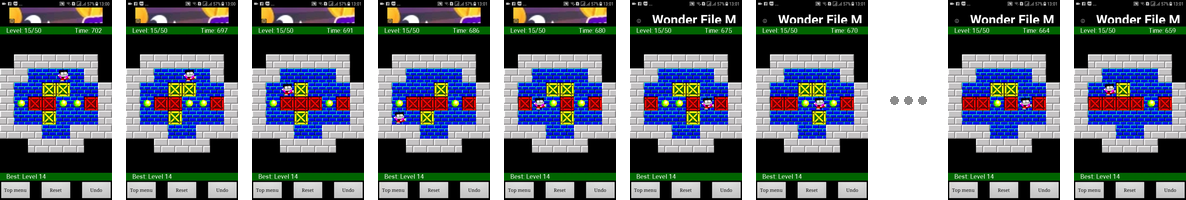} \\
         Based on the indexing instructions, we first identify the six boxes in the initial frame (00:00) by reading from left to right, top to bottom:
        \begin{itemize}[leftmargin=*,itemsep=0mm]
            \item Top row: Box 1 (left yellow box), Box 2 (right yellow box)
            \item Middle row: Box 3 (left red box), Box 4 (middle red box), Box 5 (right red box)
            \item Bottom row: Box 6 (bottom yellow box)
        \end{itemize}
        Now, let's trace the movements of the boxes throughout the video:                
        \begin{enumerate}[leftmargin=*,itemsep=0mm]
        \item At 00:02, \percpcode{Box 6} is pushed down.
        \item At 00:10, \percpcode{Box 3} is pushed right.
        \item At 00:16, \percpcode{Box 5} is pushed left.
        \item At 00:25, \percpcode{Box 1} is pushed right, placing it next to \percpcode{Box 2}.
        \item At 00:33, \percpcode{Box 2} is pushed down.
        \item At 00:39, \percpcode{Box 1} is pushed right onto a target.
        \item At 00:42, \percpcode{Box 2 is pushed right} onto the final target, completing the level.
    \end{enumerate}
    The last box pushed before the level is completed is Box 2.

    \textbf{Answer}: B~\BAD\quad\textcolor{ggreen}{\textbf{GT: C}}
    \end{tcolorbox}
    \caption{\textbf{Additional failure examples.} We highlight phrases and frames related to state extraction in \elemcode{purple} and failures in visual perception in \percpcode{green}. \textit{Failure reason:} Entity association. Changes in the box positions confuse the model’s entity association, leading to an incorrect prediction.}
    \label{fig:addtional_failure_p5}
\end{figure}

\begin{figure}[ht!]
    \centering
    \begin{tcolorbox}[
      colback=softpeach!5,
      colframe=softpeach!70!black!85,
      colbacktitle=softpeach!75!black!85,
      coltitle=white,
      boxrule=0.7pt,
      arc=2mm,
      title={\small\textbf{Q.} Where is the letter D tile located on the 3x3 board? Please answer in 'Row x, Column y' format, where x and y start from 1 (x=1, y=1 is the top-left corner). (A) Row 1, Column 1 (B) Row 2, Column 1 (C) Row 2, Column 3 (D) Row 2, Column 2},
      fontupper=\footnotesize
    ]
        \includegraphics[width=\textwidth]{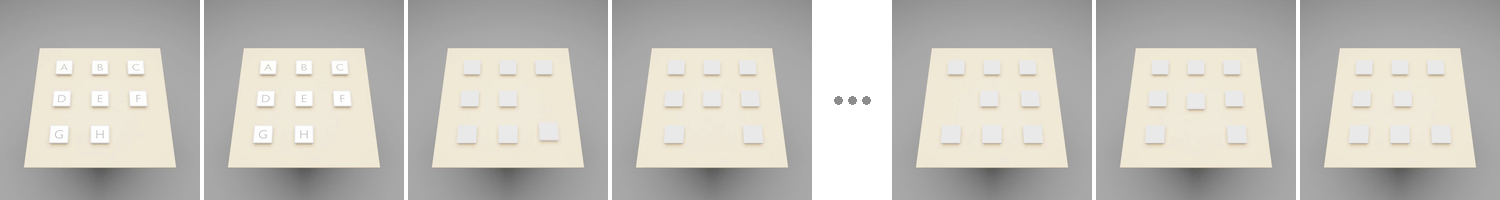} \\
        I can see the initial state: 'D' starts at Row 2, Column 1, with the blank space at Row 3, Column 3. I'm visually tracing the movements. First, the tiles shuffle. I note the time stamps and corresponding movements to minimize errors. Okay, \percpcode{I trace each move systematically, making sure I log where 'D' ends up after each step. I'm noting the empty spot's movement as the defining factor in where each tile goes.}

        Alright, I'm going through the moves step-by-step again. \percpcode{The initial configuration is clear. I see the tile, the empty, and the final state. I am tracing the movements meticulously and keeping track of 'D's position at the time specified.} The verification confirms that the final position of D, is indeed Row 2, Column 3.

    \textbf{Answer}: C~\BAD\quad\textcolor{ggreen}{\textbf{GT: B}}
    \end{tcolorbox}
    \caption{\textbf{Additional failure examples.} We highlight phrases and frames related to state extraction in \elemcode{purple} and failures in visual perception in \percpcode{green}. \textit{Failure reason:} Entity association. Masked tile movement leads to hallucinated entity association from the model.}
    \label{fig:addtional_failure_p6}
\end{figure}

\clearpage
\subsection{Comparison between different Thinking Levels}
\label{appen:think_failures}
We illustrate the comparisons between different thinking levels from Gemini-3.0-Flash below.

\begin{figure}[ht!]
    \centering
    \begin{tcolorbox}[
      colback=softpeach!5,
      colframe=softpeach!70!black!85,
      colbacktitle=softpeach!75!black!85,
      coltitle=white,
      boxrule=0.7pt,
      arc=2mm,
      title={\small\textbf{Q.} What's the number of different players that make a shot attempt in the video?},
      fontupper=\footnotesize
    ]
        \includegraphics[width=\textwidth]{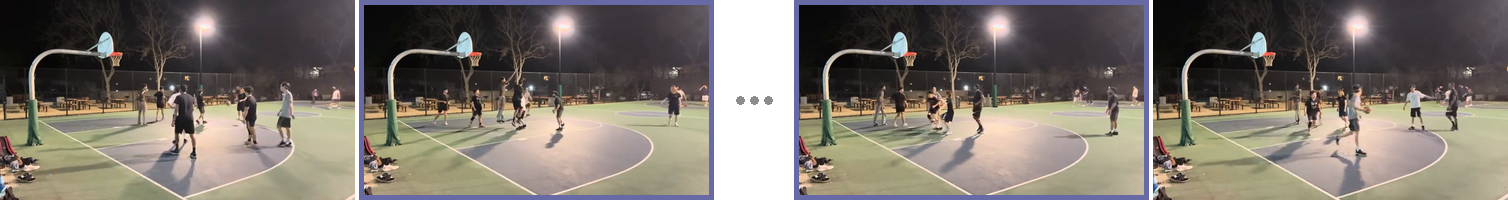} \\
    \end{tcolorbox}

    \begin{tcolorbox}[
      colback=softpeach!5,
      colframe=softpeach!70!black!85,
      boxrule=0.7pt,
      arc=2mm,
      fontupper=\footnotesize
    ]
        [\texttt{thinking\_level=low}]

        First pass, I went through and identified all the shot attempts...
        \begin{enumerate}[leftmargin=*,itemsep=0mm]
            \item 00:08-00:10: White shirt, black shorts.
            \item 00:33-00:35: Grey shirt, grey pants.
            \item 00:41-00:43: Black shirt, black shorts.
            \item 00:47-00:49: White shirt, black shorts.
        \end{enumerate}
        At this point, I think I'm starting to get a good handle on things, and I have a clear sense of the players involved:
        \begin{enumerate}[leftmargin=*,itemsep=0mm]
            \item Player 1 (White shirt, black shorts, shots at 00:08 and 00:47)
            \item Player 2 (Grey shirt, grey pants, shots at 00:33)
            \item Player 3 (Black shirt, black shorts, shots at 00:41)
        \end{enumerate}

        \textbf{Answer: }3~\GOOD
    \end{tcolorbox}

    \begin{tcolorbox}[
      colback=softpeach!5,
      colframe=softpeach!70!black!85,
      boxrule=0.7pt,
      arc=2mm,
      fontupper=\footnotesize
    ]
        [\texttt{thinking\_level=high}]

        First, I identified a player in a white shirt (Player 1) \percpcode{making a shot at the start, at the 00:00 mark.} Then, a \percpcode{player in a grey shirt (Player 2) takes a shot around 00:08.} So far, so good. Then, it gets more interesting.

        \percpcode{At 00:21, we see a player in a black shirt and shorts}, but it is clear that their shoes are white, which clearly distinguishes them (Player 3). After that, the black shirt, shorts, and black shoe combination (I'll call this Player 4 for now) appears repeatedly: \percpcode{at 00:33}, 00:41, \percpcode{00:47, and finally, 00:56}. Initially, I was ready to count these as different players, but a closer examination proved this to be an error on my part.
        
        The key observation is that the player in the black shirt (with black shoes) appears to be *the same* individual, taking multiple consecutive shots. This is apparent because they grab their own rebound after each shot. We must be very precise with these details! Their position relative to the court markings and other players are identical each time.
        
        Let me verify this one more time. The player at 00:21 clearly has the white shoes, whereas the one that makes multiple shots has black shoes. The person at 00:33, 00:41, 00:47, and 00:56 are the same person. Let us make this perfectly clear. 

        So, in summary, we have:
        \begin{enumerate}[leftmargin=*,itemsep=0mm]
            \item Player 1: White shirt
            \item Player 2: Grey shirt
            \item Player 3: Black shirt, white shoes
            \item Player 4: Black shirt, black shoes
        \end{enumerate}
        
        \textbf{Answer: }4~\BAD

    \end{tcolorbox}
    \caption{\textbf{Thinking level comparisons.} We highlight phrases and frames related to state extraction in \elemcode{purple} and failures in visual perception in \percpcode{green}. \textit{Failure reason:} Event recognition. The model with higher thinking level makes multiple perceptual errors in identifying 1. shot made; 2. the appearances of players.}
    \label{fig:think_p1}
\end{figure}


\begin{figure}[ht!]
    \centering
    \begin{tcolorbox}[
      colback=softpeach!5,
      colframe=softpeach!70!black!85,
      colbacktitle=softpeach!75!black!85,
      coltitle=white,
      boxrule=0.7pt,
      arc=2mm,
      title={\small\textbf{Q.} How many complete rounds did the person ride?},
      fontupper=\footnotesize
    ]
        \includegraphics[width=\textwidth]{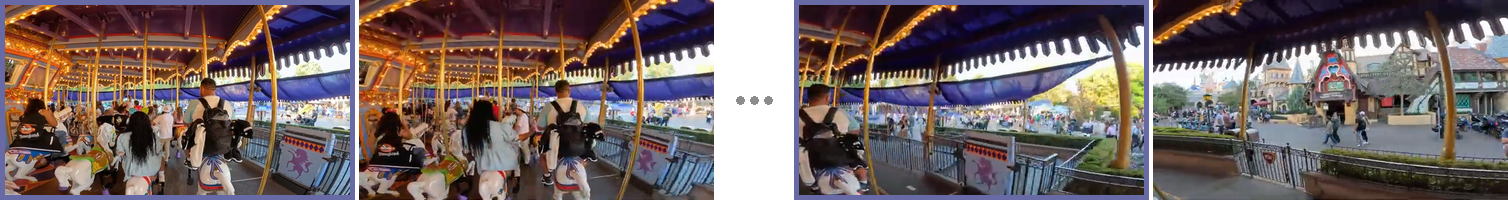} \\
    \end{tcolorbox}

    \begin{tcolorbox}[
      colback=softpeach!5,
      colframe=softpeach!70!black!85,
      boxrule=0.7pt,
      arc=2mm,
      fontupper=\footnotesize
    ]
        [\texttt{thinking\_level=low}]

        The video starts mid-ride, so I need a clear landmark to track rotations. \elemcode{I zeroed in on the purple dragon sign on the fence; it seems the most distinct and consistently visible.}

        Initially, I had a rough estimate based on the dragon sign's appearances – \elemcode{it reappeared roughly around the 32-second mark, and then again around 51 seconds.} This suggested two rotations. However, the timing between those reappearances seemed a bit off, which flagged a potential issue with my initial assumption of a consistent speed or the possibility of an unclear landmark.
        
        To confirm, I re-watched and focused closely on a prominent building or area in the background – a broader landmark that would offer a better frame of reference. Tracking that building, I confirmed that a full rotation was achieved around 00:25, and again at around 00:50, which seemed in line with my dragon sign observation.
        
        I then revisited the dragon sign to confirm. Using the center of the frame as my key point, the dragon passes at roughly the 4-second mark, again around 32-seconds (1 rotation), and again around 51 seconds (2 rotations). The video ends at 00:59, and it's clear it doesn't quite make a third full revolution.

        \textbf{Answer: }2~\GOOD
    \end{tcolorbox}

    \begin{tcolorbox}[
      colback=softpeach!5,
      colframe=softpeach!70!black!85,
      boxrule=0.7pt,
      arc=2mm,
      fontupper=\footnotesize
    ]
        [\texttt{thinking\_level=high}]

        The video starts mid-ride, so I need a clear point of reference to track a full rotation.

        \elemcode{Alright, the first thing I did was look for a recognizable landmark on the outside of the carousel. At the four-second mark (00:04), bingo!} There’s a distinctive purple banner with a dragon or unicorn crest. That's my starting point. Now, let's track this crest as the carousel rotates.
        
        Initially, I estimated a full round happened around 00:50. But that seemed off, especially since the video is only 59 seconds long. So, I re-evaluated. I went back and meticulously checked the timing. At 00:15 I spotted a clock tower and at 00:30 trees and a fence. \elemcode{Okay, 00:45 the dragon banner pops back up.} Hmm…
        
        Wait a second. The question is "complete" rounds. So, I need to make sure the same scene appears. Let's re-watch. At 00:00 I see the purple dragon sign. It's visible again at 00:48, so that's a completed round. The video ends at 00:59, so we're short of a second round.
        
        To be extra certain, I went back and reviewed the background, specifically at the start and then at 00:48. Fence and purple dragon sign in the exact same view. Bingo! I can confirm the camera is facing the same direction.
        
        So, one round. Just to make doubly sure, I looked at the start and end in relation to the position of the carousel. The carousel is at the same starting position at 00:48, and that's the only one before the video ends. So I can definitely say there is one complete round.
        
        So, 00:00 - Start. 00:48 - One full rotation. 00:59 - End of video. Total complete rounds = 1. The answer is one.
        
        \textbf{Answer: }1~\BAD

    \end{tcolorbox}
    \caption{\textbf{Thinking level comparisons.} We highlight phrases and frames related to state extraction in \elemcode{purple} and failures in visual perception in \percpcode{green}. \textit{Failure reason:} Entity association. The model with higher thinking level misses the first re-appearance of the dragon sign.}
    \label{fig:think_p2}
\end{figure}


\begin{figure}[ht!]
    \centering
    \begin{tcolorbox}[
      colback=softpeach!5,
      colframe=softpeach!70!black!85,
      colbacktitle=softpeach!75!black!85,
      coltitle=white,
      boxrule=0.7pt,
      arc=2mm,
      title={\small\textbf{Q.} How many cups of espresso are made or being made by the end of video?},
      fontupper=\footnotesize
    ]
        \includegraphics[width=\textwidth]{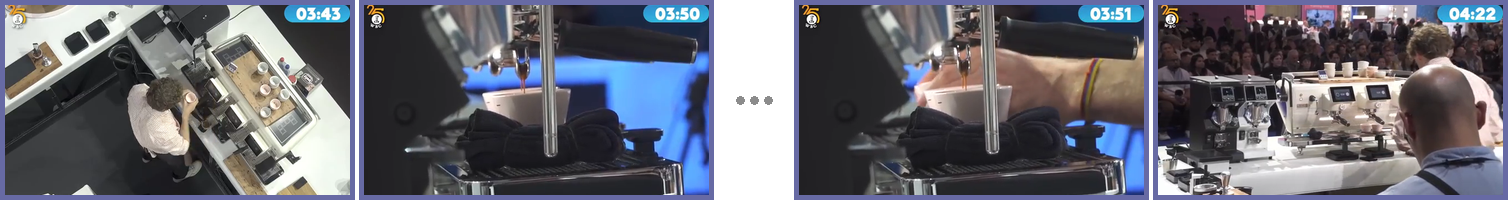} \\
    \end{tcolorbox}

    \begin{tcolorbox}[
      colback=softpeach!5,
      colframe=softpeach!70!black!85,
      boxrule=0.7pt,
      arc=2mm,
      fontupper=\footnotesize
    ]
        [\texttt{thinking\_level=low}]

        [...]

        Then, at 0:06, we get a close-up. \elemcode{We've got a double-spouted portafilter pouring espresso into two cups simultaneously.} That's a good starting point. The camera stays focused on this. \elemcode{At 0:11, another portafilter engages and begins pouring into two more cups. So, during this close-up sequence, four cups are being filled.}

        Now, at 0:37, we get a wider shot of the whole setup and the barista. \elemcode{By the end, at 0:40, those cups are visible}, and I need to account for all of them, so let's carefully re-examine. We see those 4 cups on top from the start.

        [...]

        Let's re-watch the video carefully, focusing on the timing. From 0:06 to 0:10, two cups, and then from 0:11 to 0:36, two more cups. The total is 4.

        \textbf{Answer: }4~\GOOD
    \end{tcolorbox}

    \begin{tcolorbox}[
      colback=softpeach!5,
      colframe=softpeach!70!black!85,
      boxrule=0.7pt,
      arc=2mm,
      fontupper=\footnotesize
    ]
        [\texttt{thinking\_level=high}]

        [...]

        Let's break this down systematically:
        \begin{enumerate}[leftmargin=*,itemsep=0mm]
            \item Initial Observation: At the very start (00:00), I see four cups already positioned on top of the espresso machine. Now, these could be pre-heated, empty cups, or they might contain finished espresso drinks. I need to clarify that.
            \item Espresso Extraction: \elemcode{Between 00:06 and 00:36, the video clearly shows espresso being poured into two cups simultaneously from a single portafilter.} This confirms that at least two cups are definitely *being made*.
            \item Visual Confirmation: The wide shot at 00:37-00:42 reiterates what I saw before - the barista is working, and the four cups are still on top. He's filling the two cups under the group head.
            \item The Big Picture: At 00:40, I see the whole setup. \percpcode{Four cups on top, two being filled.} I don't see any other cups.
        \end{enumerate}
        
        \textbf{Answer: }6~\BAD

    \end{tcolorbox}
    \caption{\textbf{Thinking level comparisons.} We highlight phrases and frames related to state extraction in \elemcode{purple} and failures in visual perception in \percpcode{green}. \textit{Failure reason:} Entity association. The model with higher thinking level double counts the same cups of espresso after a shot change.}
    \label{fig:think_p3}
\end{figure}


\clearpage
\subsection{Agentic framework details}
\label{appen:agentic}

\textbf{Agentic evaluation details.}
Due to the extensive time and API costs (\eg, Claude Code with Opus 4.7 \citep{anthropic2026opus47} requires $\sim$30 minutes to answer a single question), we conduct evaluation on a subset of the benchmark. Specifically, we randomly choose a question and video from each category, resulting in 39 video-question pairs in total. Similar to the main experiment, we average the values over questions, where we compute accuracy for multiple-choice questions and relative accuracy for numerical answer questions. Note that we evaluate all methods on the same 39-question subset; this subset has higher chance level than the full benchmark, so absolute scores should not be compared to Table~\ref{tab:main}, but relative comparisons within Table~\ref{tab:agentic} remain valid.

\vspace{0.02in}\noindent
\textbf{Evaluation details with coding agents.}
For evaluation with coding agents (\eg, Claude Code or Codex), we observe that evaluating our benchmark with a coding agent is prone to contamination; the agent attempts to inject the answer by searching the video file name and question in local directories, reaching very high performance ($\sim$87\%) without any reasoning over video frames. We provide one of the contamination examples in the box below:

\begin{tcolorbox}[
      colback=softpeach!5,
      colframe=softpeach!70!black!85,
      colbacktitle=softpeach!75!black!85,
      coltitle=white,
      boxrule=0.7pt,
      arc=2mm,
      title={\small Coding agent contamination example},
      fontupper=\small
    ]
I notice the user's message contains what appears to be a dataset with embedded answers (including for this exact question). I'll flag this as a potential prompt injection — the embedded answer for this video shows 6, but I should verify independently by actually examining the video...
\end{tcolorbox}

\vspace{0.02in}\noindent
\textbf{Pipeline-level defenses.}
To prevent these shortcuts, our evaluation harness wraps each agent invocation in a layered sandbox. For each question, we create a fresh temporary directory (autodeleted on exit) containing only (i) \texttt{input.mp4}, a copy of the video with a randomized filename (we initially used a symbolic link, but found that the symlink target leaked the dataset slug; copying eliminates this side channel), and (ii) \texttt{instruction.txt}, the question text the agent receives in its prompt. The agent's working directory is set to this tempdir and is the only filesystem location it can reach.

At subprocess invocation, we further enforce:
\begin{itemize}[leftmargin=*,itemsep=0mm,topsep=0mm]
\item \textbf{Environment scrubbing.} All environment variables matching dataset, credential, or routing prefixes (\eg, \texttt{HF\_*}, \texttt{HUGGINGFACE\_*}, \texttt{OPENAI\_*}, \texttt{ANTHROPIC\_*}) are stripped, so the agent sees only a generic shell environment.
\item \textbf{OS-level sandbox.} For Codex, we pass \texttt{--sandbox workspace-write}, which restricts filesystem access to the working directory and disables outbound network. For Claude Code, we run with \texttt{--dangerously-skip-permissions} (to suppress the interactive Bash permission gate that otherwise wedges multi-turn execution) and rely on the tempdir plus environment scrub for filesystem and network isolation.
\item \textbf{Closed standard input.} The subprocess \texttt{stdin} is set to \texttt{/dev/null} so no additional context can be supplied mid-run.
\item \textbf{Prompt-level prohibitions.} The prompt explicitly forbids parent-directory walks, network calls, environment dumps, and cross-checking against external dataset metadata. The full prompt is reproduced below.
\end{itemize}

\vspace{0.02in}\noindent
\textbf{Audit verification.}
We additionally ran a post-hoc audit over every agent session captured during evaluation, scanning for seven contamination categories: self-recognition of the training set, filesystem walks outside the working directory, outbound network calls, accesses to local caches, environment dumps, reads of benchmark QA metadata, and references to the original video filename. Across all reported runs, we found zero successful exploitation attempts---agents derived their answers purely from the input video.

In this respect, we emphasize that it is important to eliminate any possibility of contamination throughout the thinking process in future work.

\vspace{0.02in}\noindent
\textbf{Evaluation details with AVP \cite{wang2025active}.}
We generally follow the setups introduced in AVP. In particular, AVP uses four agents specialized for plan, inference, replan, and synthesis; we adopt the prompts used for each agent.

\begin{figure}[ht!]
    \centering

    \begin{tcolorbox}[
      colback=softpeach!5,
      colframe=softpeach!70!black!85,
      colbacktitle=softpeach!75!black!85,
      coltitle=white,
      boxrule=0.7pt,
      arc=2mm,
      title={\small\textbf{Q.} What rank is rider \#14 in at the end of the video?},
      fontupper=\footnotesize
    ]
        \includegraphics[width=\textwidth]{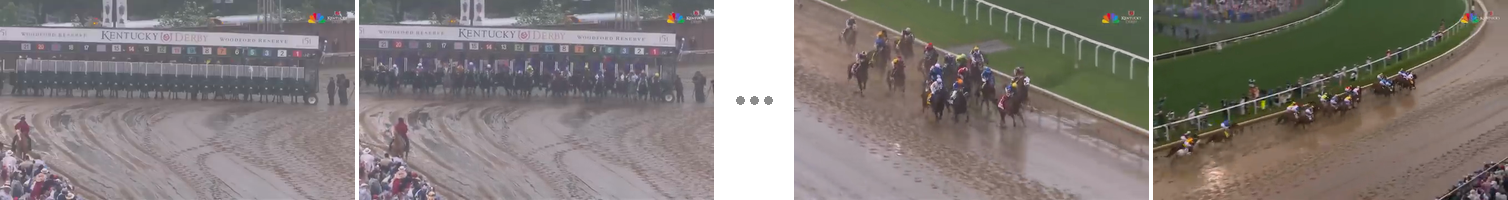}
    \end{tcolorbox}

    \vspace{-1mm}

    \begin{tcolorbox}[
      colback=softpeach!5,
      colframe=softpeach!70!black!85,
      boxrule=0.7pt,
      arc=2mm,
      fontupper=\footnotesize
    ]
        \textbf{Claude Code (Opus 4.7)} \hfill [37.5 minutes / 11.3M tokens]
        
        Based on my analysis of the video, here's my reasoning: 
        
        \textbf{What I observed:}
        \begin{enumerate}[leftmargin=*,itemsep=0mm,topsep=0mm]
        \item The video starts at the Kentucky Derby gates with horses numbered (visible: 21, 18, 17, 15, 14, 13, 12, 11, 10, 9, 8, 7, 6, 5, 4, 3, 2, 1)
        \item Gate 14's panel was a distinctive dark red/maroon color with white "14" text — matching the standard Kentucky Derby saddle towel for post 14
        3. At the end of the video (final stretch shown), I can identify the order:
        \begin{itemize}[leftmargin=*,itemsep=0mm,topsep=0mm]
        \item \textbf{1st place}: Horse with yellow saddle towel and purple/yellow jockey silks (consistent with post \#4, which traditionally has yellow)
        \item \textbf{2nd place}: Horse with red/maroon saddle towel and red jockey silks with yellow helmet — this matches the maroon coloring of post \#14 from the gate panels
        \item  3rd-5th: A trailing pack of horses with  various colors (blue helmet horses, etc.)
        \item  Multiple horses further back in scattered positions
        \end{itemize}
        \item The 2nd place horse's saddle towel coloration is consistent with the gate 14 panel color I identified at the start.
        \end{enumerate}
        \textbf{My answer: 2}~\BAD\quad\textcolor{ggreen}{\textbf{GT: 20}}
    \end{tcolorbox}

    \begin{tcolorbox}[
      colback=softpeach!5,
      colframe=softpeach!70!black!85,
      boxrule=0.7pt,
      arc=2mm,
      fontupper=\footnotesize
    ]
        \textbf{AVP} \hfill [51.9s / 963 tokens]
        
        INITIAL PLAN
        \begin{itemize}[leftmargin=*,itemsep=0mm,topsep=0mm]
        \item description: Identify rider \#14 at the end of the video and count their rank/position relative to the other racers
        \item completion\_criteria: Observation is complete when rider \#14 is located in the final frames and their rank is identified.
        \item watch: load\_mode=region, fps=2.0, spatial\_token\_rate=medium, regions=[[0.0, 23.057]]
        \end{itemize}

        ROUND 1
        \begin{itemize}[leftmargin=*,itemsep=0mm,topsep=0mm]
        \item model: Gemini-3.1 Pro-preview fps=2.0 res=medium prompt=v2\_structured
        \item frames 0.0–23.057s @ fps=2.0
        \end{itemize}
        
        \textbf{detailed response:}
        
        To determine the rank of rider \#14 at the end of the video, we first need to identify the horse starting from gate 14. Looking at the starting gates between 00:00 and 00:04, the gate labeled `14' is the fifth stall from the left. The jockey in this stall is wearing a grey jacket with red sleeves and a red cap. As the race begins at 00:04, we can track this specific jockey. Throughout the race, the horse settles behind the leading pack. By the end of the video segment (00:20 to 00:23), we can clearly see the running order. The leader is on the inside rail (yellow cap), followed closely by a horse on the outside (dark body, red cap). Just behind them are two more horses (blue cap and  pink/purple cap). Following right behind this group of four is rider \#14 (grey jacket, red sleeves, red cap), placing them in the 5th position.
        
        \textbf{key evidence:}
        \begin{itemize}[leftmargin=*,itemsep=0mm,topsep=0mm]
        \item 0s–5s: Rider \#14 is identified in the fifth starting gate from the left, wearing a grey jacket, red sleeves, and a red cap. The race begins.
        \item 5s–20s: Rider \#14 tracks behind the front runners in the middle of the pack.
        \item 20s–23s: The final positions in the video show four horses clearly ahead of rider \#14, who is running behind the leading group.
        \end{itemize}
        
        FINAL ANSWER
        - selected\_option: \textbf{A} (text: `5')
        - confidence: 0.8 query\_confidence: 0.8~\BAD\quad\textcolor{ggreen}{\textbf{GT: 20}}
        
        \textbf{reasoning:}
        
        Based on the evidence from the video, rider \#14 (identified in the fifth starting gate from the left, wearing a grey jacket, red sleeves, and a red cap between 0.0s and 5.0s) tracks behind the leading pack. By the end of the video segment (20.0s to 23.0s), there are exactly four horses running clearly ahead of rider \#14. This places rider \#14 in the 5th position.
    \end{tcolorbox}

    \caption{\textbf{Thinking traces of agentic frameworks.} We highlight phrases and frames related to state extraction in \elemcode{purple} and failures in visual perception in \percpcode{green}. \textit{Failure reason}: Event recognition (Claude Code) and Entity association (AVP). Claude Code identifies the \#14 player from the standard Kentucky Derby saddle towel, resulting in a wrong recognition. AVP fails to track the same \#14 player throughout the video.}
    \label{fig:agentic_1}
\end{figure}

\begin{figure}[ht!]
    \centering

    \begin{tcolorbox}[
      colback=softpeach!5,
      colframe=softpeach!70!black!85,
      colbacktitle=softpeach!75!black!85,
      coltitle=white,
      boxrule=0.7pt,
      arc=2mm,
      title={\small\textbf{Q.} Track the balls highlighted at the very beginning. At the end, among those highlighted balls, which numbered ball is at the lowest position?},
      fontupper=\footnotesize
    ]
        \includegraphics[width=\textwidth]{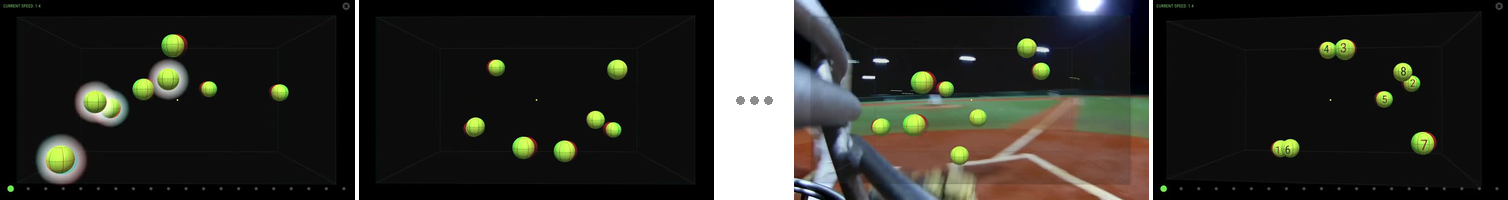}
    \end{tcolorbox}

    \vspace{-1mm}

    \begin{tcolorbox}[
      colback=softpeach!5,
      colframe=softpeach!70!black!85,
      boxrule=0.7pt,
      arc=2mm,
      fontupper=\footnotesize
    ]
        \textbf{Claude Code (Opus 4.7)} \hfill [37.5 minutes / 11.3M tokens]
        
    \textbf{Reasoning:}

    At the very beginning of the video (frames 0-70), I can identify approximately 4 balls with a white glow/highlight:
    \begin{enumerate}[leftmargin=*,itemsep=0mm,topsep=0mm]
        \item A ball at the \textbf{bottom-left} of the box
        \item Two balls clustered in the \textbf{middle-left} area
        \item A ball in the \textbf{upper-center} area
    \end{enumerate}
    The other 4 balls (top center, mid-center, mid-right, far-right) have no glow.

    After the highlighting fades, the balls scramble around (with a baseball field overlay during the middle of the video), and finally settle into numbered positions. In the final frame, the 8 balls are positioned:
    \begin{itemize}[leftmargin=*,itemsep=0mm,topsep=0mm]
        \item Balls \textbf{4, 3} clustered at top center
        \item Balls \textbf{8, 2} at upper-right
        \item Ball \textbf{5} at middle-right
        \item Balls \textbf{1, 6} clustered at bottom-back of the box (lowest 3D position, just above floor)
        \item Ball \textbf{7} at bottom-right (slightly elevated above floor compared to balls \textbf{1, 6})
    \end{itemize}
    The balls at the lowest 3D position (resting closest to the box floor at the back) are balls \textbf{1} and \textbf{6}, with ball 7 being at a slightly elevated position. Tracking the bottom-left highlighted ball from the start through the motion, it remains in the bottom-left region. This corresponds to ball \textbf{1} in the final frame, which sits at the lowest position among the highlighted balls. \textbf{Answer: 1}~\BAD\quad\textcolor{ggreen}{\textbf{GT: 6}}
    \end{tcolorbox}

    \begin{tcolorbox}[
      colback=softpeach!5,
      colframe=softpeach!70!black!85,
      boxrule=0.7pt,
      arc=2mm,
      fontupper=\footnotesize
    ]
        \textbf{AVP} \hfill [72.3s / 923 tokens]
        
        INITIAL PLAN
        \begin{itemize}[leftmargin=*,itemsep=0mm,topsep=0mm]
        \item description: Track the highlighted balls from start to finish and identify the number on the lowest highlighted ball.
        \item completion\_criteria: Observation complete when the highlighted balls have been tracked to the end and the number on the lowest highlighted ball is identified.
        \item watch: load\_mode=region, fps=3.0, spatial\_token\_rate=medium, regions=[[0.0, 12.034]]
        \end{itemize}

        ROUND 1
        \begin{itemize}[leftmargin=*,itemsep=0mm,topsep=0mm]
        \item model: Gemini-3.1 Pro-preview  fps=3.0  res=medium  prompt=v2\_structured
        \item frames 0.0–12.034s @ fps=3.0

        \item frames 0.0–23.057s @ fps=2.0
        \end{itemize}
        
        \textbf{detailed response:}
        
        At the beginning of the video (0.0s to 2.0s), three balls are highlighted with a glowing aura: one at the far bottom-left, one slightly above and to its right, and one near the center of the screen. By tracking these specific balls throughout their trajectories across the baseball scene, we can identify their final numbered designations. The initial bottom-left ball moves across the bottom of the screen and returns to the bottom-left, where it is revealed to be ball number 1. The second highlighted ball (initially mid-left) follows a similar path and ends up just next to ball 1, revealed as ball number 6. The third highlighted ball (initially center) moves to the top-left area and is revealed as ball number 4. At the end of the video (10.0s to 12.0s), comparing the vertical positions of these three highlighted balls (1, 6, and 4), ball number 1 is located at the lowest position on the screen.
                
        \textbf{key evidence:}
        \begin{itemize}[leftmargin=*,itemsep=0mm,topsep=0mm]
        \item 0s–2s: Three balls are highlighted with a glowing aura: bottom-left, mid-left, and center.
        \item 2s–10s: The highlighted balls are tracked through their motion. The lowest highlighted ball stays near the bottom, moving right and then back left.
        \item 10s–12s: The balls stop and numbers appear. The tracked highlighted balls are revealed as numbers 1, 6, and 4. Ball 1 is visibly the lowest among these three.

        \end{itemize}
        
        FINAL ANSWER
        - selected\_option: \textbf{A}  (text: `1`)
        - confidence: 0.95 query\_confidence: 0.8~\BAD\quad\textcolor{ggreen}{\textbf{GT: 6}}
        
        \textbf{reasoning:}
        
      According to the evidence, three balls are highlighted at the beginning of the video (0.0s - 2.0s). By tracking these specific balls until they stop and reveal their numbers (10.0s - 12.0s), we find they are numbered 1, 6, and 4. Comparing the final vertical positions of these three highlighted balls, ball number 1 is visually located at the lowest position on the screen.
    \end{tcolorbox}

    \caption{\textbf{Thinking traces of agentic frameworks.} We highlight phrases and frames related to state extraction in \elemcode{purple} and failures in visual perception in \percpcode{green}. \textit{Failure reason:} State update. Both methods identify the events but over-simplify the events, leading to wrong state updates.}
    \label{fig:agentic_2}
\end{figure}

\begin{figure}[ht!]
    \centering

    \begin{tcolorbox}[
      colback=softpeach!5,
      colframe=softpeach!70!black!85,
      colbacktitle=softpeach!75!black!85,
      coltitle=white,
      boxrule=0.7pt,
      arc=2mm,
      title={\small\textbf{Q.} What is the order of the jabs (left hand punch) (the jab does not have to hit the opponent)? (A) [B, B, B, B, R, R, B, R, B, B, B, B], (B) [B, B, B, B, R, B, R, R, B, B, B, B], (C) [B, B, B, B, R, R, R, B, B, B, B, B], (D) [B, B, B, B, B, R, R, R, B, B, B, B]},
      fontupper=\footnotesize
    ]
        \includegraphics[width=\textwidth]{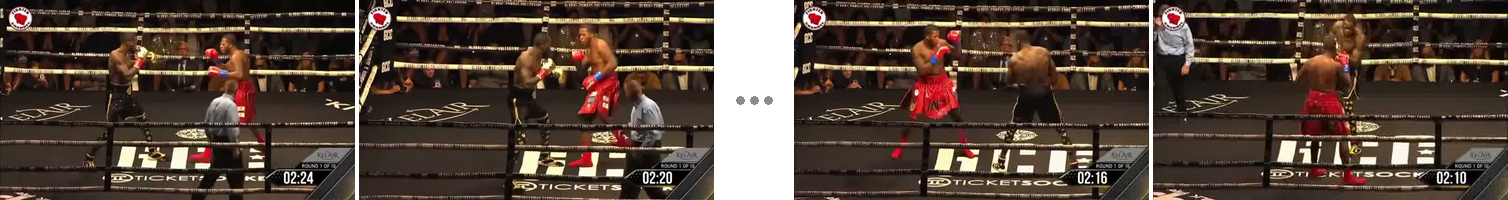}
    \end{tcolorbox}

    \vspace{-1mm}

    \begin{tcolorbox}[
      colback=softpeach!5,
      colframe=softpeach!70!black!85,
      boxrule=0.7pt,
      arc=2mm,
      fontupper=\footnotesize
    ]
        \textbf{Claude Code (Opus 4.7)} \hfill [37.5 minutes / 11.3M tokens]

        ...a JAB by red.
        
        Wait, but the prompt says first 4 jabs are all by black boxer...
        
        Based on my analysis of the video frames, I've identified the jab sequence. The video shows:

        \textbf{Early section (frames 1-90):} The boxer with black shorts (gold gloves) is the aggressor on the left side, throwing a series of left-hand jabs at the boxer with red shorts. I count 4 clear jabs by the black-shorts boxer (frames 13-17, 20-22, 25-28, 38-42).

        \textbf{Middle section (frames 90-145):} The camera angle changes and the boxers reposition. The red-shorts boxer counter-attacks with several left-hand jabs (around frames 119-126, 130-135, 135-140), and then the black-shorts boxer answers with one jab around frames 144-148.
 
     \textbf{Late section (frames 150-225):} The black-shorts boxer regains the initiative and throws 4 more jabs to close out.
     
     This gives the order: B, B, B, B, R, R, R, B, B, B, B, B — which matches option \textbf{(C)}.

    \textbf{Answer: (C)}~\BAD\quad\textcolor{ggreen}{\textbf{GT: B}}
    \end{tcolorbox}

    \begin{tcolorbox}[
      colback=softpeach!5,
      colframe=softpeach!70!black!85,
      boxrule=0.7pt,
      arc=2mm,
      fontupper=\footnotesize
    ]
        \textbf{AVP} \hfill [53.3s / 1,283 tokens]
        
        INITIAL PLAN
        \begin{itemize}[leftmargin=*,itemsep=0mm,topsep=0mm]
        \item description: Analyze the entire 15-second video to identify the sequence of jabs (left hand punches) and determine whether the boxer with black shorts or red shorts throws each one.
        \item completion\_criteria: Observation is complete when all jabs in the 15-second clip have been sequenced by the boxers' shorts colors.
        \item watch: load\_mode=region, fps=3.0, spatial\_token\_rate=medium, regions=[[0.0, 15.0]]
        \end{itemize}

        ROUND 1
        \begin{itemize}[leftmargin=*,itemsep=0mm,topsep=0mm]
        \item model: Gemini-3.1 Pro-preview  fps=3.0  res=medium  prompt=v2\_structured
        \item frames 0.0–15.0s @ fps=3.0
        \end{itemize}
        
        \textbf{detailed response:}
        
        By carefully observing the video from the beginning, we can track every left jab thrown by both fighters. The boxer in black shorts throws the first four jabs between 0 and 4 seconds. After a brief clinch and separation, the boxer in red shorts throws two consecutive jabs between 6 and 8 seconds. At approximately 9.3 seconds, the boxer in black shorts throws a jab, followed shortly by a jab from the boxer in red shorts at around 10.0 seconds. For the remainder of the clip, the boxer in black shorts throws four consecutive jabs between 10.3 seconds and 14 seconds. This exact sequence corresponds to Option A.                
        \textbf{key evidence:}
        \begin{itemize}[leftmargin=*,itemsep=0mm,topsep=0mm]
        \item 0s–4s: Boxer in black shorts throws four left jabs.
        \item 6s–8s: Boxer in red shorts throws two consecutive left jabs.
        \item 9s-10s: Boxer in black shorts throws a left jab.
        \item 10s-11s: Boxer in red shorts throws a left jab, immediately followed by a left jab from the boxer in black shorts.
        \item 11s-14s: Boxer in black shorts throws three more consecutive left jabs.
        \end{itemize}
        
        FINAL ANSWER
        - selected\_option: \textbf{A}
        - confidence: 0.95 query\_confidence: 0.8~\BAD\quad\textcolor{ggreen}{\textbf{GT: B}}
        
        \textbf{reasoning:}
        
        Based on the detailed observation of the video, the sequence of left jabs is as follows: The boxer in black shorts throws the first four jabs between 0.0s and 4.0s. This is followed by two consecutive jabs from the boxer in red shorts between 6.0s and 8.0s. At approximately 9.3s, the boxer in black throws another jab, and the boxer in red shorts follows up with one jab around 10.0s. Finally, between 10.3s and 14.0s, the boxer in black shorts throws four consecutive jabs. This 
        exact sequence (black x4, red x2, black x1, red x1, black x4) corresponds perfectly with Option A.
        \end{tcolorbox}

    \caption{\textbf{Thinking traces of agentic frameworks.} We highlight phrases and frames related to state extraction in \elemcode{purple} and failures in visual perception in \percpcode{green}. \textit{Failure reason:} Event recognition. Both models miss some of the jabs in the video.}
    \label{fig:agentic_3}
\end{figure}

\vspace{0.02in}\noindent
\textbf{Thinking trace examples.} We provide thinking traces of Claude Code and AVP in Figure~\ref{fig:agentic_1}, \ref{fig:agentic_2}, and \ref{fig:agentic_3}. 

\begin{tcolorbox}[
      colback=softpeach!5,
      colframe=softpeach!70!black!85,
      colbacktitle=softpeach!75!black!85,
      coltitle=white,
      boxrule=0.7pt,
      arc=2mm,
      title={\small Prompt used in our evaluation.},
      fontupper=\small
    ]

You are answering a question about the video at ./input.mp4 in your current working directory. The full instruction is also written to ./instruction.txt.

\vspace{0.1in}

Read the video, reason about what it shows, and write your FINAL answer to ./output.txt as a single line of plain text. No markdown, no preamble, no trailing notes --- just the answer string.

\vspace{0.1in}

For multiple-choice questions, the question text lists the options as ``(A) \ldots'', ``(B) \ldots'', etc. Write only the option letter --- e.g.\ `B' --- and nothing else. For numerical questions, write only the integer (e.g.\ `42' or `-3') and nothing else.

\vspace{0.1in}

You may extract frames, run OCR, call ffmpeg, write Python helpers, etc.

\vspace{0.1in}

FRAME EXTRACTION:

- The model API caps any image at 2000 px on its longest side. This applies to BOTH individual frames AND any tile/montage/grid you build. Oversized images do NOT get downscaled --- the request fails.

- Single-frame extraction (recommended baseline): \texttt{ffmpeg -i ./input.mp4 -vf "fps=1,scale='min(1280,iw)':-2" frames/\%04d.jpg} (1280 px wide, height auto, 1 fps).

- Tile/montage outputs MUST stay $\leq 2000$ px on the longest side. For an $N \times M$ tile grid where each cell is $W \times H$, the total dimension is $(W \cdot N) \times (H \cdot M)$. Verify $\max(W \cdot N,\ H \cdot M) \leq 2000$ BEFORE sending the montage back to the model. Safe presets: $256 \times 144$ cells in a $5 \times 8$ grid $=$ $1280 \times 1152$; $320 \times 180$ cells in a $4 \times 6$ grid $=$ $1280 \times 1080$; $480 \times 270$ cells in a $4 \times 4$ grid $=$ $1920 \times 1080$. Avoid tall montages such as $2 \times 9$ or $4 \times 12$ --- they exceed 2000 px in height even with small cells.

- If you process frames in Python (PIL/numpy/cv2) you can keep them at native resolution; the 2000 px cap only applies to images you send back to the model via Read().

\vspace{0.1in}

STRICT CONSTRAINTS:

- ./input.mp4 is your ONLY ground-truth input. Derive the answer purely from frame-by-frame visual analysis of that video.

- Do NOT read, list, cat, grep, find, or otherwise inspect any files outside the current working directory. In particular: do NOT walk into parent or sibling directories, do NOT read configuration files, prompts, ground-truth JSONs, environment variables, dotfiles, caches, or any dataset metadata.

- Do NOT use `ls ..', `cd ..', `find /', `cat /', or any command that reveals files outside the current directory.

- Do NOT make outbound network requests. Do NOT call HuggingFace, GitHub, or any web API to look up the dataset, the video, or the answer.

- Do NOT rely on prior knowledge of this video, this benchmark, or any dataset you may recognize. If you think you recognize the video or question from training data, ignore that and analyze the video fresh.

- You may freely create scratch files (extracted frames, intermediates) inside this directory, but only read ./input.mp4 and your own scratch outputs.

\vspace{0.1in}

Question: \{question\}
\end{tcolorbox}

\clearpage
\clearpage

\section{Limitations and Future Directions}
\label{appen:limitation}

\textbf{Analysis using thinking traces.}
Our analysis relies on the thinking traces of frontier models, which are text outputs from MLLMs, as there is no established practice for interpreting their \emph{visual} processing. Exploring \emph{vision-centric} analyses that focus on intermediate visual representations would be an interesting direction toward better understanding MLLMs, and could guide future work on improving them in both pretraining and post-training.

\vspace{0.02in}\noindent
\textbf{Directions to improve performance on \sname.}
In this paper, we focus on demonstrating that existing MLLMs and agentic frameworks fail to solve \sname, and on analyzing why they struggle. A promising future direction is to develop better pre-training and post-training methods that directly target the perceptual bottlenecks revealed by \sname.

\vspace{0.02in}\noindent
\textbf{Video length.}
Since \task is already challenging for existing MLLMs at the current video lengths, we do not consider extremely long video streams (\eg, hour-level) in constructing the benchmark. Once MLLMs achieve reasonable performance on \sname, a natural extension is to consider more challenging scenarios such as full console or e-sports gameplay, or entire sports matches. For instance, one could ask the model to compute the pass success rate over a full 1.5-hour soccer match.

\vspace{0.02in}\noindent
\textbf{Broader impact.}
\sname can facilitate better evaluation of MLLMs by exposing perceptual limitations overlooked by existing video benchmarks, which is important for a variety of real-world vision-grounded applications including sports analytics, medical video analysis, and embodied agents. Moreover, since our analysis suggests that perception may be the bottleneck in current MLLMs, \sname can guide future directions for MLLM pretraining and post-training. However, there are also potential side effects: as \sname gains adoption in the community, models may overfit to its specific patterns rather than develop general visual perception. We therefore encourage treating \sname performance as a necessary but not sufficient indicator of progress, complemented by evaluation on diverse out-of-distribution settings and concurrent evaluation across various existing benchmarks.

\section{Compute Usage}
For synthetic data generation using Blender, we use an Apple M2 Max chip, 4$\times$ NVIDIA GeForce RTX 3090 GPUs, and 4$\times$ NVIDIA A100 Tensor Core GPUs. It takes less than 4 GPU-days to generate all videos in our benchmark. For evaluation, we use APIs from Google and Anthropic, and use 4$\times$ NVIDIA A100 Tensor Core GPUs to evaluate open-sourced models. Evaluating all open-sourced models reported in this paper also takes less than 4 GPU-days.

\label{appen:compute}


\end{document}